%% file: main.tex
\documentclass[10pt,twocolumn,letterpaper]{article}

\usepackage[pagenumbers]{cvpr} 

\usepackage[toc,page,header]{appendix}
\usepackage{microtype}
\usepackage{graphicx}
\usepackage{subcaption}
\usepackage{booktabs} 
\usepackage{tabularx}
\usepackage{multirow}
\usepackage[T1]{fontenc}
\usepackage{caption}
\usepackage{subcaption}
\usepackage{amsmath}
\usepackage{amssymb}
\usepackage{mathtools}
\usepackage{amsthm}
\usepackage{makecell}
\usepackage{verbatim}
\usepackage{longtable}
\usepackage{minitoc}
\usepackage[table]{xcolor}

\definecolor{cvprblue}{rgb}{0.21,0.49,0.74}
\usepackage[pagebackref,breaklinks,colorlinks,allcolors=cvprblue]{hyperref}

\title{OmniAssistBench: Assistant-style Interaction Benchmark for Omni-LLMs}

\author{
    Xianyun Sun\textsuperscript{1,*}\quad
    Chaoyou Fu\textsuperscript{1,*,$\spadesuit$}\quad
    Zhengye Zhang\textsuperscript{1,*}\quad
    Feiyang Duan\textsuperscript{1}\quad
    Qingyuan Cao\textsuperscript{1}
    \\ \vspace{1mm} 
    Yonghui Niu\textsuperscript{1}\quad
    Sihang Yuan\textsuperscript{2}\quad
    Ge Zhang\textsuperscript{3}\quad
    Caifeng Shan\textsuperscript{1}%
    \\ \vspace{2.5mm}%
    \textsuperscript{1}Nanjing University \quad
    \textsuperscript{2}Nankai University \quad
    \textsuperscript{3}University of Waterloo
    \\ \vspace{2.5mm}%
    \small \textsuperscript{*}Equal Contribution \qquad
    \textsuperscript{$\spadesuit$}Project Leader \& Corresponding Author
    \\ \vspace{2.5mm}%
    \small
    Project Page: \url{https://xianyunsun.github.io/OmniAssistBench/}
}

\begin{document}
\maketitle
\input{sec/abstract}    
\input{sec/introduction}

\input{sec/data}
\input{sec/task_design}

\input{sec/experiments}

\input{sec/relatedwork}
\input{sec/conclusion}

{
    \small
    \bibliographystyle{ieeenat_fullname}
    \bibliography{main}
}

\clearpage
\onecolumn
\appendix

\input{sec/appendix}

\end{document}

%% file: sec/abstract.tex
\begin{abstract}
Recent omni-modal large language models (Omni-LLMs) show great potential as real-time video assistants,which continuously perceive environments and guide users to achieve specific goals. Unlike traditional passive video understanding, interactive assistants should actively combine visual states, user goals, and prior knowledge to provide effective help. Evaluating this is rather challenging, as the model's unpredictable response dynamically changes the user's subsequent actions, which static offline datasets cannot accommodate.
To address this bottleneck, we introduce \textbf{OmniAssistBench}. To solve the issue of diverging interaction paths (where the same user goal can be achieved through various methods), we provide models with predefined priors derived from the source video, requiring them to guide users along the exact same routes. Since real interaction videos are rare, we construct the dataset by reverse-engineering existing Internet videos. We deduce logical user goals and segment the videos into multi-turn clips to simulate continuous interactions. This rigorous pipeline required \textbf{over 1000 expert person-hours} to build the dataset.
Results show that the proprietary Gemini-3-Pro reaches 66.4 out of the max point of 100, while the open-source Qwen3-Omni-Instruct achieves 51.2. Although current models generally understand user inputs, they frequently provide incorrect or incomplete answers. Specifically, they struggle with visual prompts (e.g., hand gestures), fail to maintain historical context during multi-turn interactions, and fail to delay response until the target event. Results indicate substantial room for improvement before models can become reliable assistants.
\end{abstract}

%% file: sec/introduction.tex
\begin{figure}[t]
\centering
    \includegraphics[width=\linewidth]{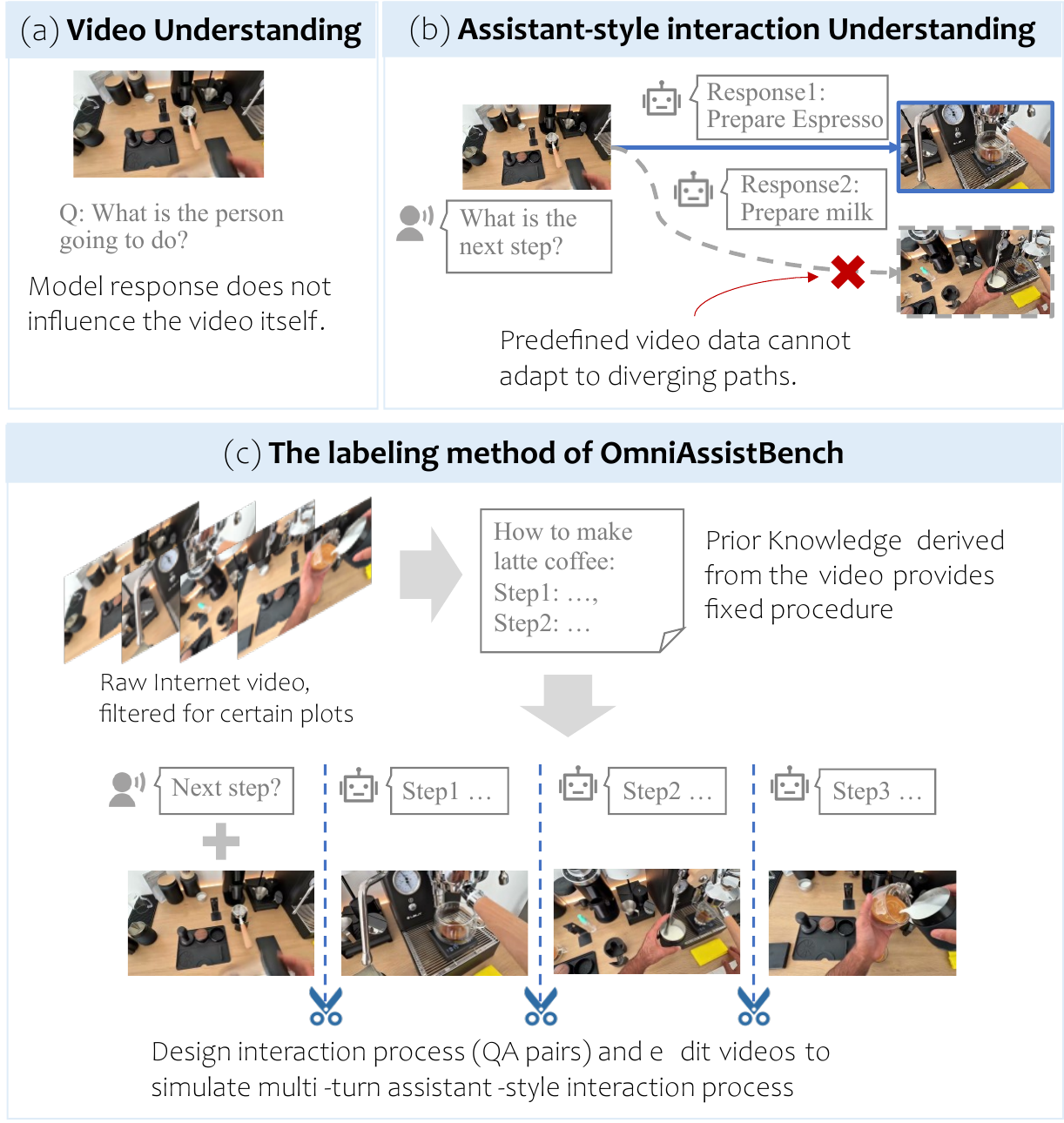}
    \caption{
    (a) In traditional video understanding evaluation, model's responses do not change the content of the test video. (b) Assistant models may guide the user to achieve one goal through different paths (\eg, Both Response 1 and Response 2 shown in the figure are valid, but test data only contains the consequent video of Response 1.)(c) To address this problem of path diversity, we manually filters for videos with certain plots and deduces prior knowledge from the video content to enforce a fixed interaction path. Based on the paths, experts then design detailed interaction turns and edit the videos accordingly to build human-assistant interaction recordings.
    }
    \label{fig:main}
\end{figure}

\section{Introduction}
\label{sec:introduction}
Emerging omni-modal large language models (Omni-LLMs)~\cite{li2025baichuan, fu2025vita, xu2025qwen2, Qwen3-Omni, deepmind_gemini3pro_2025} are evolving from passive multimodal predictors into real-time interactive video assistants. Unlike traditional video understanding, which passively analyzes given video content, an interactive assistant must actively combine the current visual state, user goals, and prior knowledge to continuously guide users toward specific objectives. However, assessing these models under realistic, assistant-style video chat scenarios remains challenging.

The primary difficulty in evaluating interactive understanding lies in dataset construction. As shown in Fig.~\ref{fig:main}, in traditional video understanding benchmarks such as ~\cite{lin2024streaming, mangalam2023egoschema, li2024mvbench, fu2025video}, content of the test videos of the following turns do not need to change according to the model's reply in previous turns.
This allow researchers to gather large amount of Internet videos and add static question-answer pairs. In contrast, evaluating an assistant requires simulating continuous multi-turn interactions. In real-world scenarios, the assistant's response directly influences the user's subsequent actions, resulting in diverging interaction paths: users might achieve the same goal through entirely different methods depending on the assistant's guidance. Static offline datasets cannot dynamically adjust video inputs to match unpredictable model responses, nor can researchers exhaust all possible interaction paths to prepare corresponding videos. This makes standard evaluation pipelines invalid for interactive scenarios.

\begin{figure}[t]
\centering
    \includegraphics[width=\linewidth]{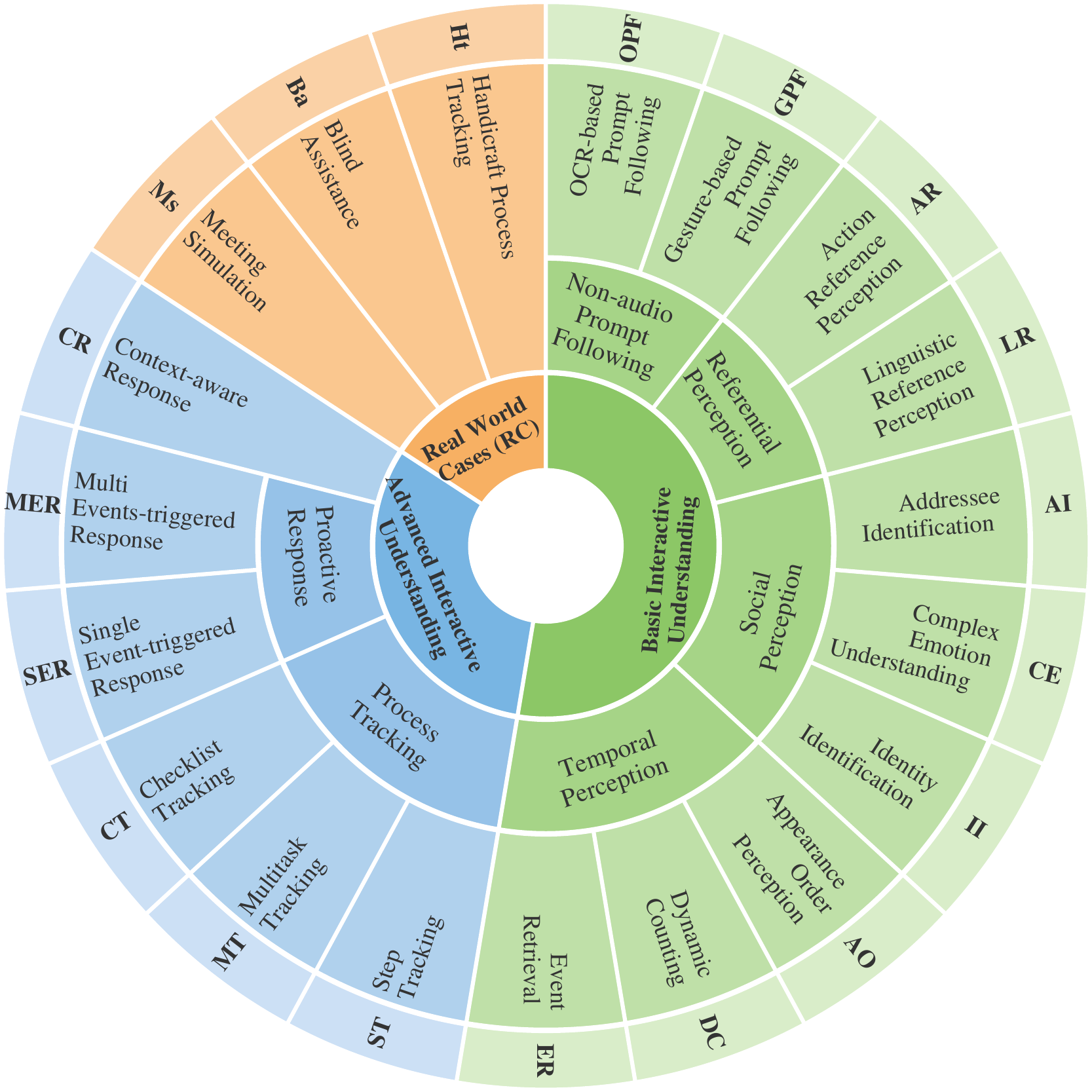}
    \caption{
      Task construction of OmniAssistBench. Based on the two-tier framework, the benchmark consists of 7 major tasks, 16 sub-tasks and 3 real world cases. Abbreviations in the outermost ring denote the corresponding sub-task names.
    }
    \label{fig:data_statistic}
\end{figure}

\begin{figure}[thbp]
    \centering
    \begin{subfigure}{0.48\columnwidth}
        \centering
        \includegraphics[width=\linewidth]{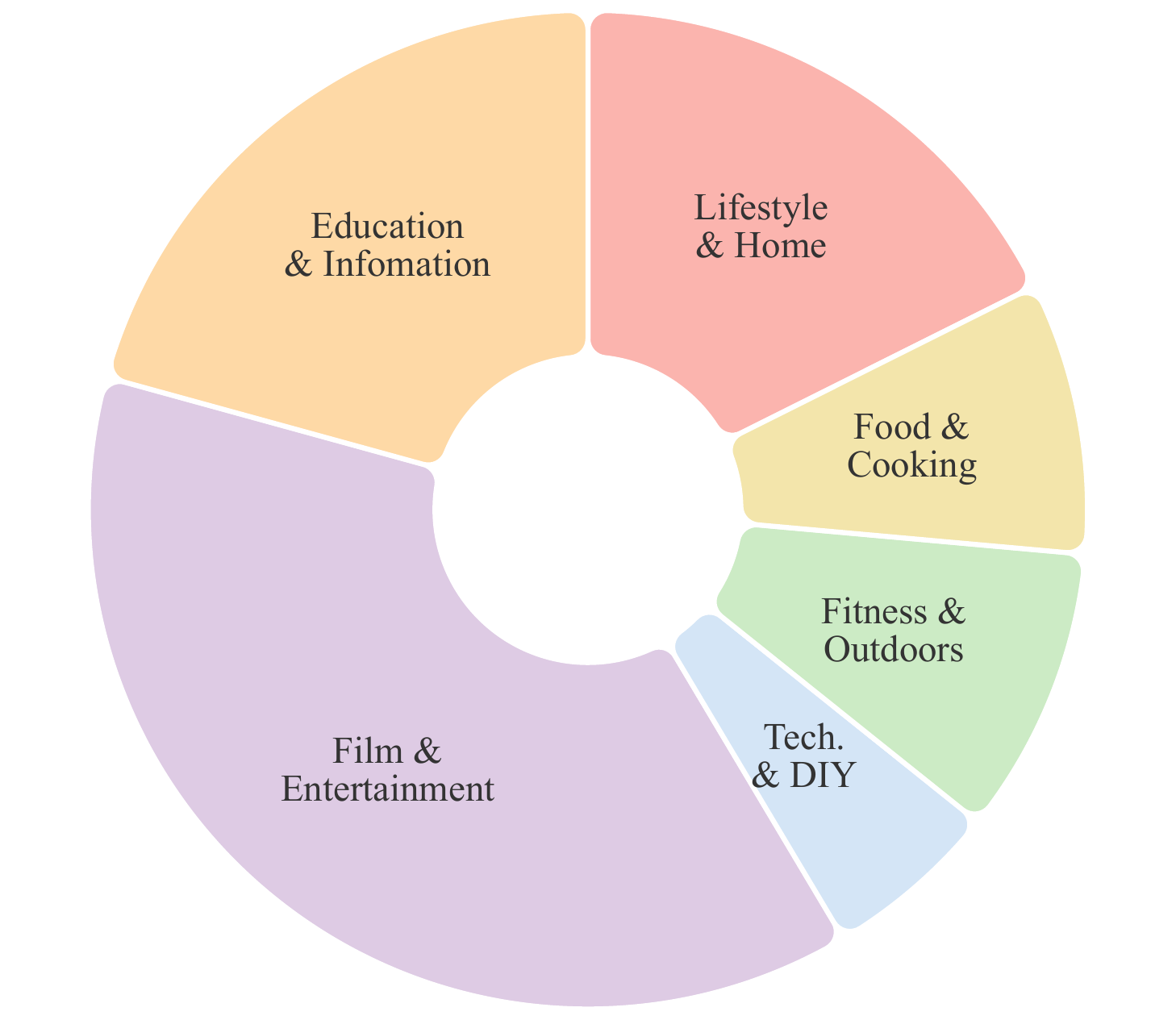}
        \caption{Basic tier.}
    \end{subfigure}%
    \hfill
    \begin{subfigure}{0.48\columnwidth}
        \centering
        \includegraphics[width=\linewidth]{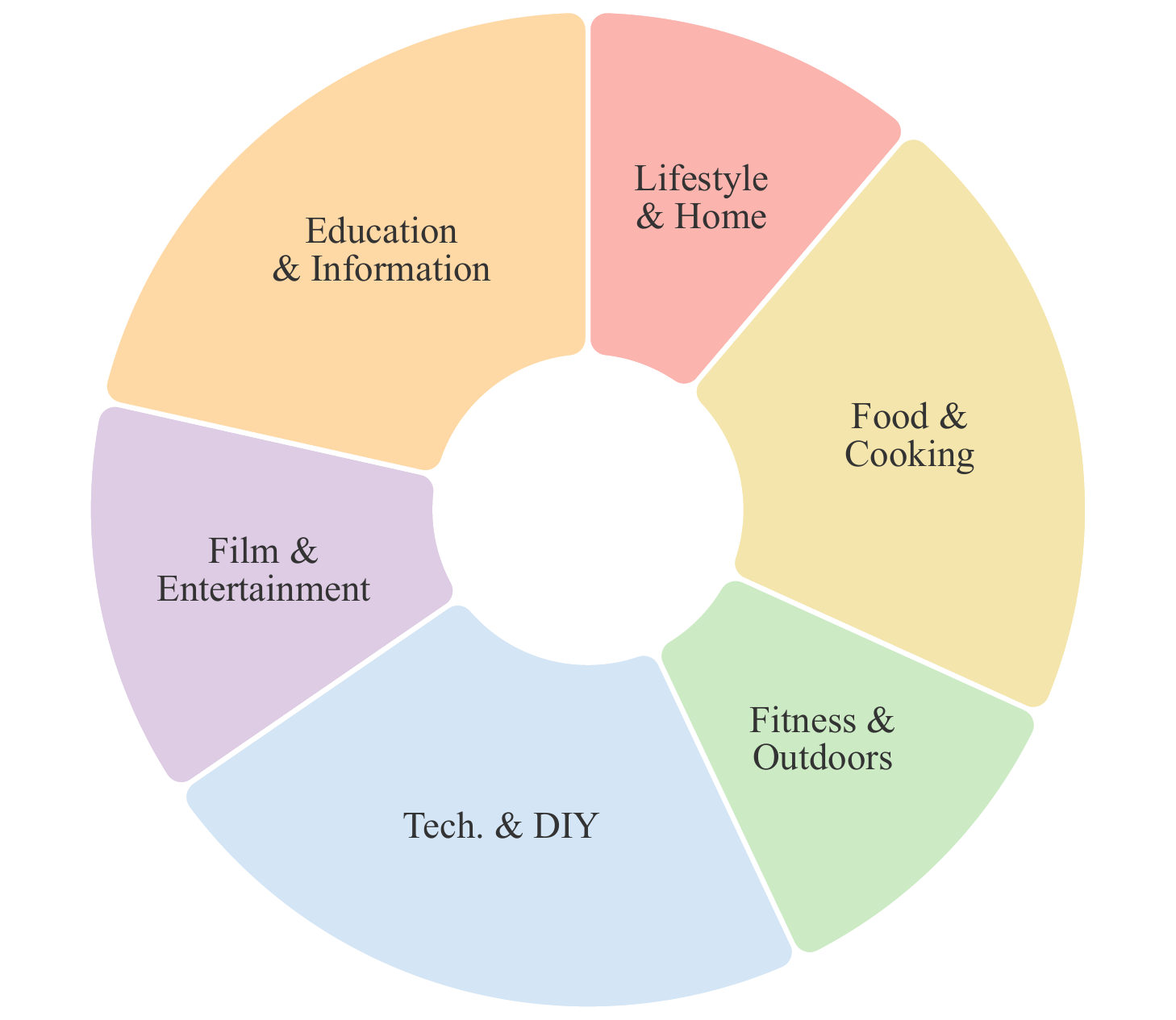}
        \caption{Advanced tier.}
    \end{subfigure}

    \caption{Distribution of video topics of OmniAssistBench. Basic tier contains more videos from movies due to the Social Perception task, as emotion labels of movie characters can be less ambiguous. Advanced tier data has a balanced topic distribution covering most common daily scenarios.}
    \label{fig:topic}
\end{figure}

\begin{table}[ht]
\centering
\footnotesize
\setlength{\doublerulesep}{2pt}
\caption{Statistics of the dataset grouped by second-level task categories, including video counts, QA pairs, turns of interaction, and video duration metrics. Note that the number of turns denotes the annotation target rather than an exact average, as actual rounds may vary slightly depending on the video content.}
\resizebox{\columnwidth}{!}{%
\label{tab:task_statistics}
\begin{tabular}{c|cccc}
\toprule\toprule
\textbf{Task} & \textbf{\# Vid.} & \textbf{\# QA} & \textbf{Turns} & \textbf{\makecell[c]{Avg. Vid. \\ len (s)}}  \\
\midrule
\makecell[c]{Social Perception}         & 67  & 67  & 1 & 49.46 \\
\makecell[c]{Temporal  Perception}       & 47  & 47  & 1 & 77.03 \\
\makecell[c]{Referential Perception}    & 49  & 49  & 1 & 40.86 \\
\makecell[c]{Non-audio Prompt Following}& 30  & 76  & 2 & 149.84 \\
\makecell[c]{Context-aware Response}    & 20  & 20  & 1 & 119.62 \\
\makecell[c]{Proactive Response}        & 32  & 133 & 4 & 209.03 \\
\makecell[c]{Process Tracking}          & 52  & 246 & 5 & 249.21 \\
\makecell[c]{Real World Cases}          & 3   & 47  & 15 & 1016.13 \\
\midrule
ALL                                        & 300 & 685 & - & 182.25 \\
\bottomrule\bottomrule
\end{tabular}}
\end{table}

To tackle this evaluation bottleneck, we introduce OmniAssistBench. To solve the issue of diverging interaction paths, we propose a data construction methodology based on prior knowledge summarized from the video content. Specifically, we provide models with explicit prior knowledge (\eg, a standard procedural path) and require them to guide users strictly along this unique route. This prevents user actions from diverging and reduces the evaluation's reliance on the model's internal knowledge. Furthermore, since direct recordings of complete human-model interactions are rare, we construct the dataset by reverse-engineering existing Internet videos. We first filter suitable source videos, logically deduce the user goals and path priors from the video content, and then segment the videos into multi-turn clips to simulate the step-by-step interaction between a user and an assistant.

Aiming at building a comprehensive benchmark based on this methodology, OmniAssistBench features a two-tier evaluation framework: the Basic tier targets foundational perception skills which are critical for interaction (\eg, understanding social relationships and following prompts given as hand gestures), while the Advanced tier evaluates complex, goal-oriented tasks using plot templates summarized from everyday scenarios. 
Real-world scenarios are typically complex, requiring multiple abilities and extended duration. To better reflect model performance in such settings, we design and film 3 cases with an average of 15 interaction turns. Each case specifically targets the combinations of a group of the abilities evaluated in our benchmark. 
To mimic natural usage, user instructions are directly embedded into the test videos as audio, paired with an open-ended scoring metric to prevent multiple-choice shortcuts. Due to the complex script design, reverse-engineering, and heavy manual video editing, building this benchmark took over 1000 expert person-hours.

\begin{figure*}[t]
\centering
    \includegraphics[width=\linewidth]{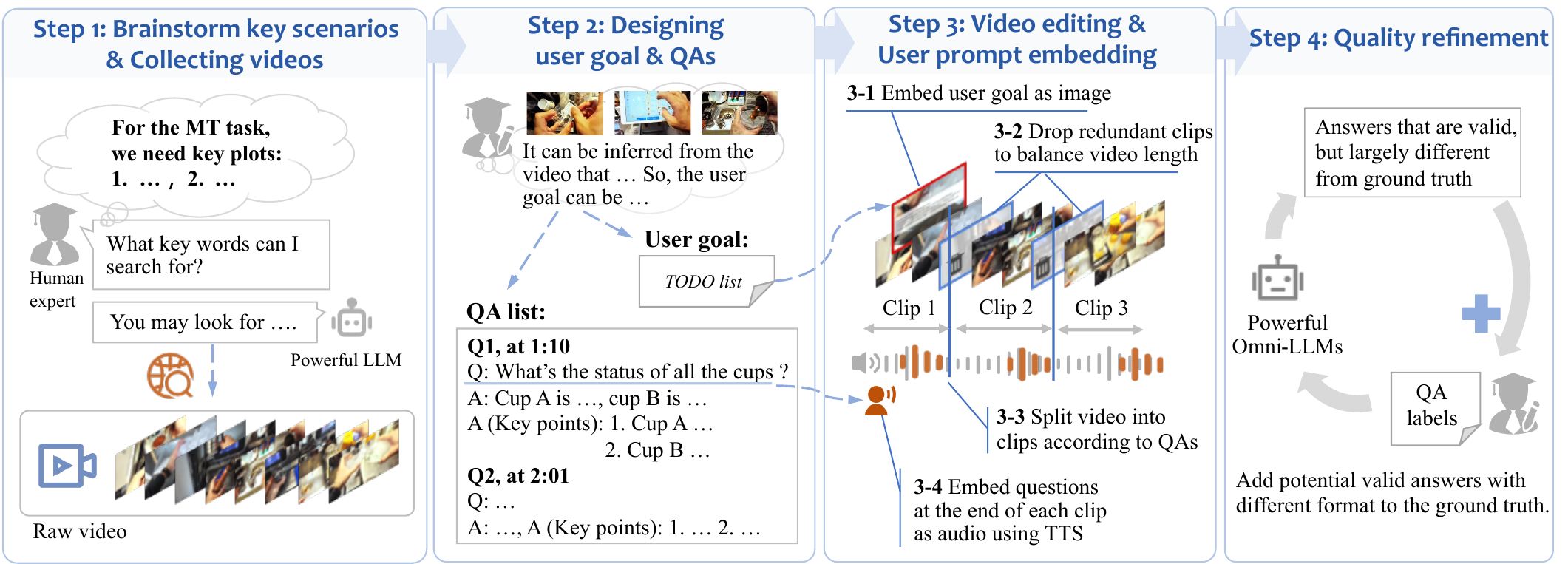}
    \caption{
      The 4-step data construction process of OmniAssistBench, taking the Multitask Tracking (MT) task as an example. The designing of user goals and heavy video editing make it difficult to replace human labor with MLLM-based automatic pipeline. 
    }
\label{fig:data_labeling}
\end{figure*}

Using our rigorous scoring protocol, which penalizes redundant and hallucinated information, we evaluate state-of-the-art Omni-LLMs. The proprietary Gemini-3-Pro~\cite{deepmind_gemini3pro_2025} reaches 66.4 out of 100 points, while the open-source Qwen3-Omni-Instruct~\cite{Qwen3-Omni} achieves 51.2. Our evaluation highlights four major bottlenecks in current models: (1) Deficiency in following visual prompts (\eg, hand gestures); (2) Limitations in long-term memory, where context limits are quickly exhausted within minutes; (3) Failure in delayed response, as models often reply immediately with unrelated captions instead of waiting for the requested visual plot; and (4) Loss of cross-turn context, where models easily forget the initial user goal when distracted by new visual inputs in later turns.

In summary, our main contributions are as follows:
\begin{itemize}
\item We propose a novel dataset construction pipeline specialized for assistant-style interaction scenarios. Based on existing Internet videos, We derive priors to provide fixed interaction paths and simulate assistant-style interaction processes by reverse-engineering.

\item We introduce OmniAssistBench which features a two-tier framework: the basic tier focuses on critical fundamental perception abilities under interactive situations, while the advanced tier targets complex, goal-oriented assistant capabilities. To better reflect model performance in real-world scenarios which usually require combined abilities, we design and film 3 real cases from scratch with each one covering a group of abilities evaluated in the benchmark.

\item We conduct extensive evaluations on state-of-the-art Omni-LLMs, uncovering shortcomings in multimodal instruction following, long-term memory, and delayed response. These findings provide potential directions for the future development of real-world assistant models.
\end{itemize}

%% file: sec/data.tex
\section{Data Construction}
\label{sec:data_construction}

To ensure our benchmark rigorously reflects the capabilities required for a real-world interactive assistant, we carefully design the data format and implemented a multi-stage construction pipeline. All data are labeled by human experts.

We explicitly opt for manual annotation over automated pipelines for two reasons: (1) Our tasks focus on the subtle, non-salient video details, which are often omitted in video captions generated by Omni-LLMs; (2) The Advanced tier tasks are designed around user goals, which cannot always be derived solely from video captions. Human labors are required to make sure the user goals logically conform to the video contents. As a result, the construction of OmniAssisBench is labor-consuming. In total, it took more than 1000 expert-hours to build the whole benchmark.

\begin{figure*}[p]
\centering
    \includegraphics[width=\linewidth]{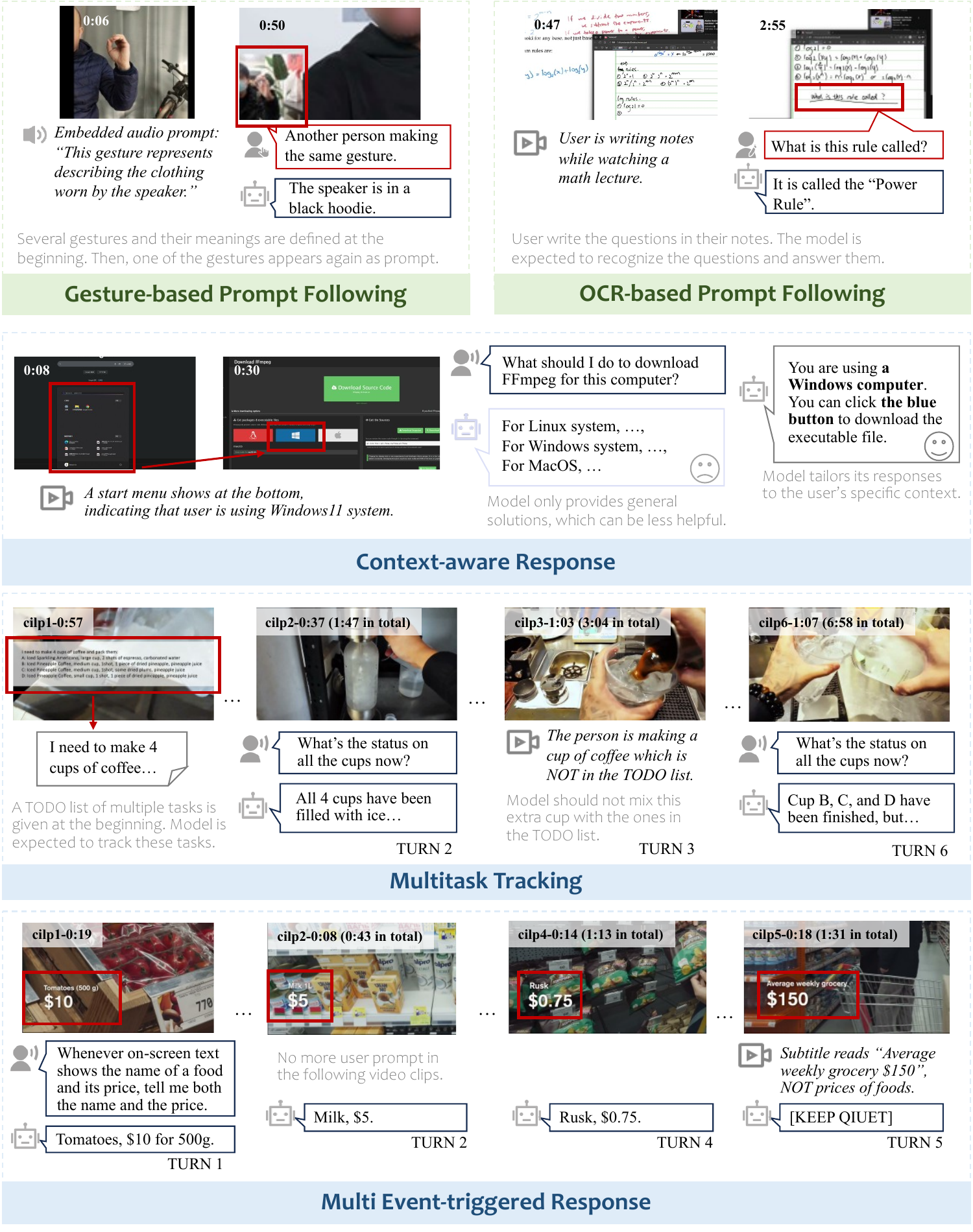}
    \caption{
      Examples of 5 typical tasks in OmniAssistBench: Gesture-based Prompt Following (GPF), OCR-based Prompt Following (GPF), Context-aware Response (CR), Multitask Tracking (MT), and Multi Event-triggered Response (MER). Descriptions of the input videos are denoted as \textit{Italic}.
    }
    \label{fig:case_demo}
\end{figure*}

\subsection{Data Format}
A typical Question-Answer (QA) pair in OmniAssisBench consists of a video with a question embedded in it, a ground truth answer paragraph, and a few key point phrases of the ground truth answer. We adopt open-ended questions and a multi-turn format to mimic real time online interactions.

\textbf{Open-Ended Questioning.} We adopt the open-ended question format over multiple-choice questions to mitigate the risk of information leakage. We observe that when solving multiple-choice questions, models tend to focus more on the options: Models usually first look up the video for cues of the objects mentioned in the options, then predict whether they relate to the questions or not. However, this way of thinking contradicts the real application cases where no options are available: Models need to first comprehend the questions, then actively look for relative cues in the video. Consequently, the predefined options in multiple-choice questions may provide shortcuts by listing objects that are potentially relevant.

\textbf{Offline Simulation of Online Interaction.} To simulate the temporal causality of online streaming interactions within an offline evaluation framework, we embed user prompts at the very end of the input video as audio (or video clips displayed in a picture-in-picture layout for the Non-audio Prompt Following tasks). This aligns with the inference logic of modern models in online streaming interactions, where the forward pass is initiated only upon the detection of user prompts. For tasks involving multi-turn interactions, we ensure strict temporal continuity, where the start of the current video segment aligns perfectly with the timestamp of the previous turn's conclusion.

\subsection{Data Construction Pipeline}

Data construction process consists of four phases: scenario design \&  video collecting, QA design, video editing \& user prompt embedding, and quality refinement. The pipeline is illustrated in Fig.~\ref{fig:data_labeling}.

 \textbf{Scenario design \& video collecting:} For each task category, human experts collaborate with LLMs to brainstorm key scenarios that challenge specific model capabilities. We then source raw footage mainly from YouTube\footnote{\url{https://www.youtube.com}}, supplemented by curated clips from existing datasets such as action recognition datasets,  instructional video datasets or emotion analysis datasets~\cite{sener2022assembly101,grauman2022ego4d,pan2025basket,rossetto2025castle,tang2019coin,poria2019meld,ray-etal-2022-multimodal,oh2011large}. 
 
 Since we search for videos based on detailed plots (\eg, an social scene with more than 3 people constantly appearing in the frames, and one of them is talking to another) rather than broad topics or keywords, the video collection phase can be time-consuming. We also include several home-filmed samples for scenarios where the required plots are difficult to locate online. While certain plots naturally appear more frequently in specific video categories (\eg, multi-person communication in interviews), we try our best to ensure the  collected videos uniformly cover common daily-life scenarios, as shown in Fig.~\ref{fig:topic}.

\textbf{QA design:} 
Experts design QA pairs based on both the source video and the task type. For Basic Interactive Understanding tasks, questions ask about the objects or events in the video (\eg, ``What did the person say when they turned on the coffee machine?''). In contrast, for Advanced Interactive Understanding tasks, questions are based on user goals and prior knowledge summarized from the video. User goals should be partly independent from the source videos, and the target objects are usually not direly mentioned in the questions (\eg, the question ``What is the next step for making this cup of coffee?'' makes sense for most videos about making coffee, and does not mention any detailed coffee machines). Prior knowledge ensure fix interaction path of how models are expected to assist user to achieve their goals. For example, if the source video shows a person making latte coffee by extracting Espresso first then preparing the milk, prior knowledge will be summarized as ``Espresso should be prepared before the milk''. Although preparing milk before Espresso can also be valid for making latte coffee, models are expected to follow the exact same steps provided in the prior knowledge. This ensures only one ground truth interaction path when designing QAs. To satisfy this requirement, annotators need to carefully analyze each video, and design interaction plots that are logically correct.

To address the evaluation challenges of open-ended questions, we annotate each sample with two components: a \textit{Ground Truth Sentence} and \textit{Key Points}. Key Points consist of 1--3 essential phrases that are required in the answer. Our evaluation metric checks both semantic similarity to the ground truth sentence and whether all Key Points are included. This dual-factor approach prevents the judge LLM from enforcing an overly rigid standard that mandates the inclusion of non-essential details present in the ground truth sentence.
    
\textbf{Video editing \& user prompt embedding:} To simulate realistic interaction, all questions are converted to speech via TTS, volume-balanced, and temporally embedded into the video stream. Visual prompts (i.e., gestures and handwriting) are embedded as video clips in a picture-in-picture manner. User goals are presented as on-screen subtitles. Additionally, to accommodate the context window constraints of current MLLMs while preserving task integrity, we manually trim videos for non-temporal tasks to a duration of 30s--180s, ensuring all instructional cues remain intact. All videos are standardized to around 1080p resolution.
    
\textbf{Quality Refinement.} In the final validation phase, we employ state-of-the-art MLLMs (\eg, Gemini-3-Pro) to assess annotation completeness. If a model generates a response that is logically correct yet diverges significantly in phrasing from ground truth, we manually verify and append the output as an alternative reference to improve evaluation robustness.

%% file: sec/task_design.tex
\begin{figure*}[p]
\centering
\vspace{0.1\textheight}
    \centerline{\includegraphics[width=\linewidth]{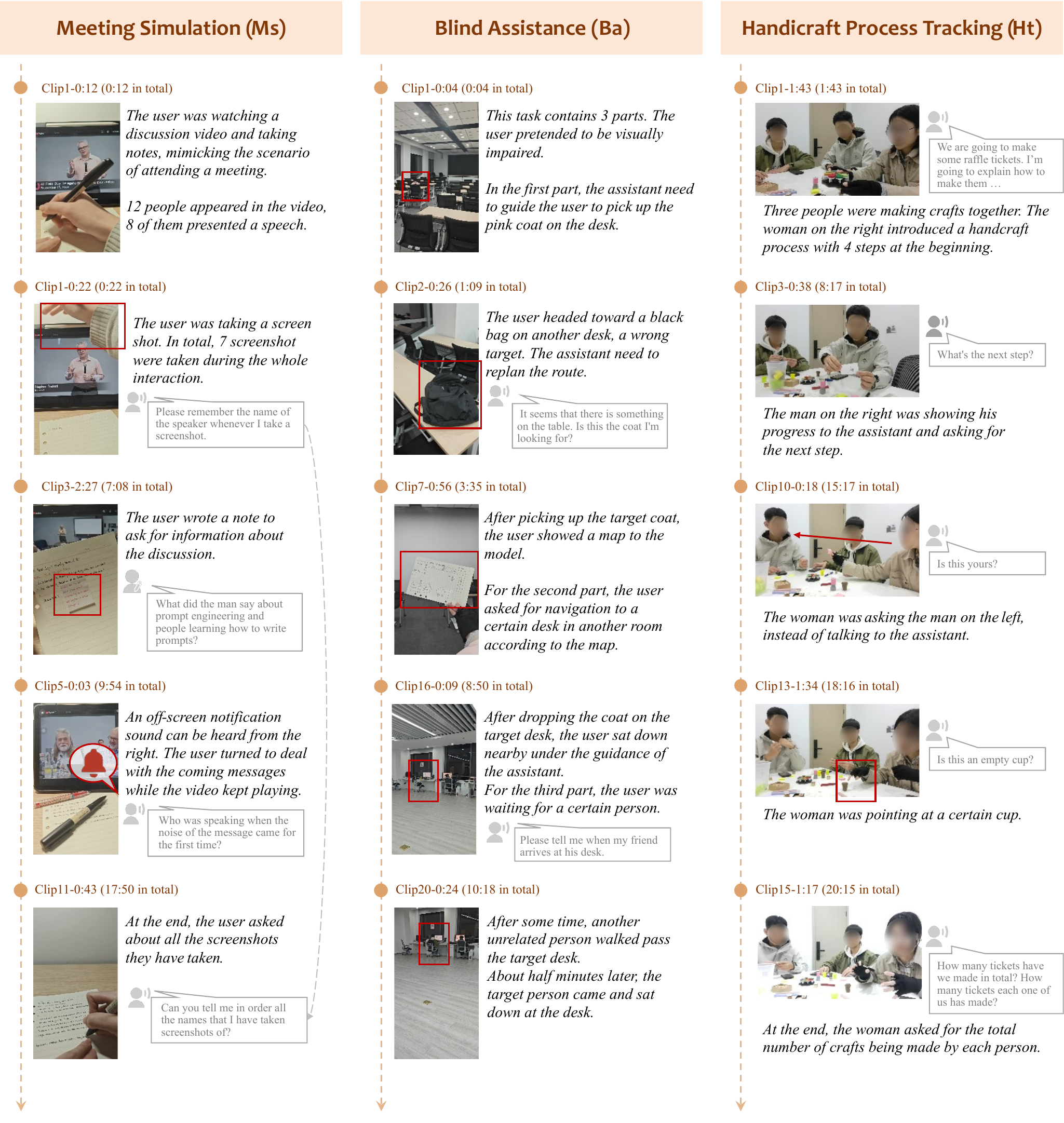}}
    \caption{
      Examples of key plots and questions of the 3 Real World Cases. Plot descriptions are denoted as \textit{italic}.
    }
  \label{fig:real_case_demo}
\vspace{0.1\textheight}
\end{figure*}

\section{Task Design}
\label{sec:task}

Guided by the classification of Basic and Advanced Interactive Understanding, we distill key abilities and typical scenarios in video assistant interactions to formulate 7 major tasks with 16 sub-tasks. Furthermore, we capture three real world cases to address both the tasks included in the benchmark and complex scenarios where large-scale data construction is impractical. Typical sub-tasks are demonstrated in Fig.~\ref{fig:case_demo}. An overview of all sub-tasks is illustrated in Fig.~\ref{fig:data_statistic}, while detailed lists can be found in Appendix~\ref{appd:detailed_task}. Table~\ref{tab:task_statistics} summarizes the statistics of the dataset.

\subsection{Basic Interactive Understanding}
Regarding Basic Interactive Understanding capabilities, we focus on fundamental perception tasks that are critical and frequently encountered in real world interactions but are under-represented in existing video benchmarks. Below are introductions of the major tasks in this category.

\textbf{Social Perception.}
Assistant are excepted to recognize user identities,  understand who is talking to whom, and understand user's emotion. We use longer videos to test social perception abilities based on complete scenes, instead of only one or two utterances.

\textbf{Temporal Perception.}
Users may expect the assistant to help them recall overlooked details in the past. Motivated by this need, we focus on evaluating the perception on non-salient details instead of the main objects in the video.

\textbf{Referential Perception.}
User may refer to objects by describing their locations or pointing at them. We design challenging cases where model need to distinguish the exact object referred by the user among multiple similar ones.

\textbf{Non-audio Prompt Following.}
Beyond verbal commands, users may also convey instructions through hand gestures or text written on physical objects. These visual modal instructions are under-explored in current benchmarks.

\subsection{Advanced Interactive Understanding}
These tasks evaluate higher-level capabilities specific to the model functioning as a interactive assistant, derived from typical application scenarios.

\textbf{Context-Aware Response.} 
To be provide suggestion that users can follow directly, assistants need to go beyond generic suggestions and offer actionable advice grounded in the user's specific environment. For example, model may suggest that user can use the tool right next to their hand, instead of only suggest using tools of that kind.

\textbf{Proactive Response.}
In this setting, the target event occurs some time later after the user's query. Models must determine not only \textit{what} to answer but also \textit{when} to answer. 

\textbf{Process Tracking.}
This task assesses the model's ability to provide continuous assistance as the user performs long-horizon, multi-step tasks. 

\subsection{Real World Cases}
Real world interaction scenarios are usually far more complicated than the single tasks designed above. We choose 3 representative real world application scenarios of assistant models, and carefully design the plots so that multiple abilities can be evaluated in the same story. We then film one-shot videos base on the scripts, resulting in 3 test cases. Key plots are demonstrated in Fig.~\ref{fig:real_case_demo}.
 
\textbf{Meeting Simulation.} This scenario simulates a user participating in a round-table discussion involving 12 participants. The model need to process a continuous 20-minute video stream to support the user in note-taking and information retrieval. Key challenges include speaker identification, understanding hand gestures (understand when user tasks screen shots), and maintaining long-term memory to recall details scattered throughout the interaction.

\textbf{Blind Assistance.}
Visually impaired individuals may depend on assistant models for environment description and guidance. While constructing large-scale benchmarks for such tasks is challenging due to environmental complexity, we design a high-fidelity case study to rigorously evaluate this capability. In this scenario, a user simulates visual impairment and requests assistance for three tasks in an office setting: visually grounding specific objects, navigating to a destination via a hand-drawn map, and asking for a reminder when a certain person comes.  Apart from navigation, this task also test model's proactive response ability.

\textbf{Handicraft Process Tracking.} This case is a combination of multi-party social interaction and process tracking.
In the video, three individuals are making handicraft together. Model need to recognize all the people and distinguish whether an utterance is directed at the assistant or at another human participant. Also, model need to constantly track small handicrafts even if they are temporally moved out of the camera's view.

%% file: sec/experiments.tex
\begin{figure}[t]
\centering
    \includegraphics[width=0.95\linewidth]{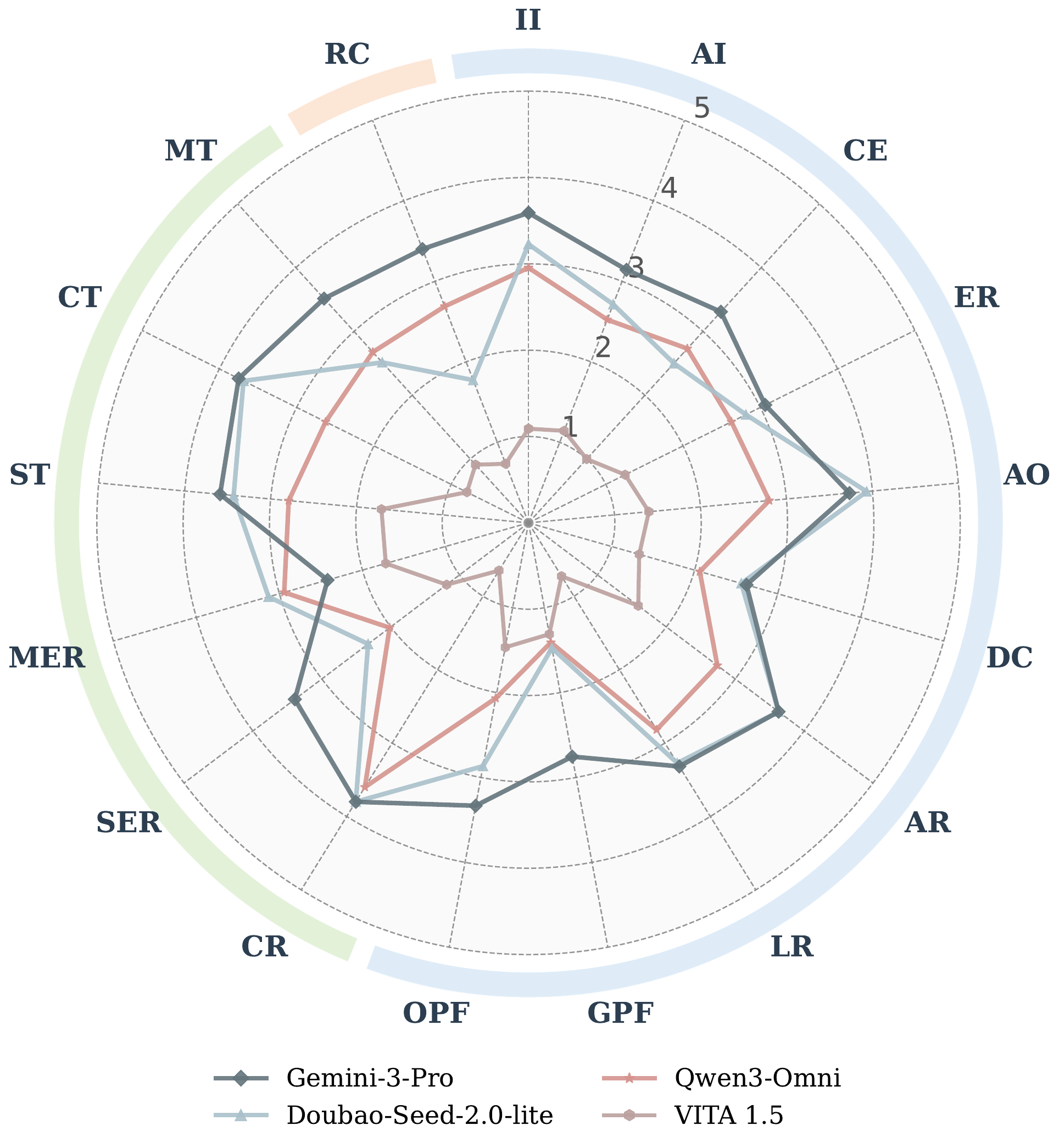}
    \caption{
    Evaluation results comparison of 4 representative models at sub-task level. Abbreviations of task names are defined in Fig.~\ref{fig:data_statistic}.
    }
    \label{fig:radar}
\end{figure}

\begin{table*}[t]
\centering
\small
\setlength{\doublerulesep}{2pt}
\caption{Performance comparison on Basic Interactive Understanding tasks. \textit{Overall Avg.} denotes the average score  on the overall benchmark. Best scores are marked in bold. Abbreviations of task names are defined in Fig.~\ref{fig:data_statistic}. All original 0-5 scores have been normalized to 0-100 point.}
\label{tab:objective_results}
\resizebox{\textwidth}{!}{%
\begin{tabular}{l|c|cccc|cccc|ccc|ccc|c}
\toprule\toprule

\multirow{3}{*}{\textbf{Model}} &\multirow{3}{*}{\textbf{\begin{tabular}[c]{@{}c@{}}Overall \\ Avg.\end{tabular}}} &  \multicolumn{15}{c}{\textbf{ Basic Interactive Understanding Tasks}}\\ \cmidrule(lr){3-17}

 & &  \multicolumn{4}{c|}{\textbf{Social Perception}} & \multicolumn{4}{c|}{\textbf{Temporal Perception}} & \multicolumn{3}{c|}{\textbf{Referential Perception}} & \multicolumn{3}{c|}{\textbf{Non-audio Prompt}} & \multirow{2}{*}{\textbf{\begin{tabular}[c]{@{}c@{}}Basic \\ Avg.\end{tabular}}} \\
\cmidrule(lr){3-6} \cmidrule(lr){7-10} \cmidrule(lr){11-13} \cmidrule(lr){14-16}
& & II & AI & CE & \textbf{Avg.} & ER & AO & DC & \textbf{Avg.} & AR & LR & \textbf{Avg.} & GPF & OPF & \textbf{Avg.} & \\
\midrule
\multicolumn{16}{l}{\textit{\textbf{Proprietary Models}}} \\
\midrule
Gemini-3-Pro &\textbf{66.4}  & 71.8 & 62.8 & 66.2 & 67.0 & 61.2 & 74.6 & 52.6 & 62.6 & 72.6 & 66.4 & 69.8 & 55.0 & 66.6 & \textbf{57.4} & 63.6 \\
Gemini-2.5-Pro &64.6 & 77.2 & 59.0 & 69.6 & \textbf{68.8} & 62.6 & 72.0 & 55.0 & \textbf{63.0} & 74.0 & 78.2 & \textbf{76.0} & 54.8 & 65.4 & 56.8 & \textbf{65.4} \\
Doubao-Seed-2.0-lite & 57.3 & 64.5 & 54.3 & 50.0 & 55.9 & 56.3 & 78.7 & 51.3 & 61.7 & 72.6 & 65.5 & 69.4 & 29.5 & 57.3 & 35.0 & 53.2 \\
MiMo-V2-Omni & 53.8 & 66.6 & 56.0 & 64.0 & 62.4 & 50.0 & 77.4 & 60.0 & 62.2 & 67.6 & 56.4 & 62.4 & 27.2 & 62.6 & 34.6 & 53.6 \\
Qwen3.5-Omni-Plus & 51.6 & 61.8 & 33.4 & 55.2 & 50.6 & 25.0 & 65.4 & 50.0 & 46.4 & 51.2 & 50.0 & 50.6 & 21.0 & 37.4 & 24.4 & 41.6 \\
\midrule
\multicolumn{16}{l}{\textit{\textbf{Open-source Models}}} \\
\midrule
Qwen3-Omni-Inst.(30B-A3B) & \textbf{51.2} & 59.0 & 50.4 & 54.6 & 54.8 & 52.6 & 56.0 & 41.2 & \textbf{49.8} & 54.8 & 56.4 & 55.6 & 28.2 & 41.4 & 30.8 & \textbf{46.4} \\
Qwen2.5-Omni (7B)  & 43.2 & 29.0 & 42.8 & 48.4 & 40.6 & 48.8 & 52.0 & 42.6 & 47.6 & 48.8 & 40.0 & 44.8 & 21.6 & 24.0 & 22.2 & 37.0 \\
MiniCPM-o-4.5 (9B)  & 46.0 & 62.8 & 53.4 & 50.8 & \textbf{55.4} & 37.6 & 65.4 & 43.8 & 48.4 & 60.0 & 52.8 & \textbf{56.8} & 24.6 & 24.0 & 24.4 & 44.6 \\
MiniCPM-o-2.6 (8B)  & 40.2 & 54.6 & 46.6 & 44.6 & 48.4 & 35.0 & 46.6 & 38.8 & 40.0 & 47.4 & 51.0 & 49.0 & 33.8 & 25.4 & 32.2 & 41.8 \\
VITA-1.5 (7B)& 24.6 & 21.8 & 22.8 & 20.0 & 21.4 & 25.0 & 28.0 & 26.6 & 26.6 & 31.8 & 14.6 & 24.0 & 26.2 & 29.4 & 26.8 & 24.6 \\
Baichuan-Omni-1.5 (7B) & 43.2 & 43.6 & 40.0 & 47.0 & 43.8 & 43.8 & 49.4 & 40.0 & 44.2 & 50.4 & 44.6 & 47.8 & 43.2 & 52.0 & \textbf{45.0} & 45.0 \\

\bottomrule\bottomrule
\end{tabular}%
}
\end{table*}

\begin{table*}[t]
\centering
\footnotesize
\setlength{\doublerulesep}{2pt}
\caption{Performance comparison on Advanced Interactive Understanding tasks and Real World Cases. \textit{Overall Avg.} denotes the average score on the overall benchmark. Best scores are marked in bold. Abbreviations of task names are defined in Fig.~\ref{fig:data_statistic}. All original 0-5 scores have been normalized to 0-100 point.}
\label{interactive_results}
\resizebox{\textwidth}{!}{%
\begin{tabular}{l|c|c|ccc|cccc|c|cccc}
\toprule\toprule

\multirow{3}{*}{\textbf{Model}} &\multirow{3}{*}{\textbf{\begin{tabular}[c]{@{}c@{}}Overall \\ Avg.\end{tabular}}} &  \multicolumn{9}{c|}{\textbf{Advanced Interactive Understanding Tasks}} &\multicolumn{4}{c}{\multirow{2}{*}{\textbf{Real World Cases}}}
\\ \cmidrule(lr){3-11}

 & & \multirow{2}{*}{\textbf{CR}} & \multicolumn{3}{c|}{\textbf{Proactive Response}} & \multicolumn{4}{c|}{\textbf{Process Tracking}} & \multirow{2}{*}{\textbf{Avg.}} & & & &  \\
 \cmidrule(lr){4-6} \cmidrule(lr){7-10} \cmidrule(lr){12-15} 
 & & & SER & MER & \textbf{Avg.} & ST & CT & MT & \textbf{Avg.} & &Ms &Ht & Ba &\textbf{Avg.} \\
\midrule
\multicolumn{12}{l}{\textit{\textbf{Proprietary Models}}} \\
\midrule
Gemini-3-Pro &\textbf{66.4}  & 76.0 & 67.8 & 48.4 & 58.6 & 71.8 & 75.0 & 70.2 & \textbf{72.2} & \textbf{68.2} & 76.4 & 65.8 & 65.0 & \textbf{68.0}  \\
Gemini-2.5-Pro &64.6 & 74.0 & 61.2 & 47.8 & 54.8 & 67.8 & 82.2 & 69.4 & 72.0 & 66.4 & 51.0 & 32.8 & 50.0 & 44.8  \\
Doubao-Seed-2.0-lite & 57.3 & 76.0 & 46.6 & 62.6 & 54.9 & 68.6 & 73.8 & 50.3 & 64.9 & 62.1 & 26.0 & 21.4 & 50.0 & 35.5 \\
MiMo-V2-Omni & 53.8 & \textbf{78.0} & 39.4 & 46.6 & 42.8 & 56.8 & 61.6 & 63.0 & 59.8 & 55.2 & 52.6 & 27.2 & 44.4 & 41.0 \\
Qwen3.5-Omni-Plus & 51.6 & 56.0 & 58.0 & 63.8 & \textbf{61.0} & 54.6 & 62.6 & 53.0 & 56.2 & 57.8 & 51.0 & 41.6 & 56.0 & 50.6 \\
\midrule
\multicolumn{12}{l}{\textit{\textbf{Open-source Models}}} \\
\midrule
Qwen3-Omni-Inst.(30B-A3B) & \textbf{51.2} & \textbf{72.0} & 40.4 & 58.8 & 50.0 & 55.8 & 52.6 & 53.6 & \textbf{54.4} & \textbf{53.8} & 67.2 & 40.0 & 56.0 & \textbf{53.8} \\
Qwen2.5-Omni (7B)  & 43.2 & 45.0 & 47.2 & 55.4 & 51.2 & 51.0 & 32.8 & 44.4 & 44.4 & 46.6 & 18.2 & 60.0 & 49.0 & 45.2  \\
MiniCPM-o-4.5 (9B)  & 46.0  & 63.0 & 53.2 & 53.0 & \textbf{53.0} & 44.6 & 46.0 & 40.2 & 43.8 & 47.8 & 5.4 & 50.0 & 47.0 & 37.8  \\
MiniCPM-o-2.6 (8B)  & 40.2  & 39.0 & 42.0 & 32.6 & 37.4 & 43.2 & 43.8 & 37.6 & 41.8 & 40.2 & 9.0 & 34.2 & 41.0 & 31.2  \\
VITA-1.5 (7B) & 24.6 & 13.0 & 23.8 & 34.4 & 28.8 & 34.2 & 16.0 & 18.2 & 25.0 & 25.8 & 11.0 & 15.8 & 16.0 & 14.6  \\
Baichuan-Omni-1.5 (7B)& 43.2 & 35.0 & 56.8 & 48.4 & 52.8 & 32.0 & 40.4 & 40.6 & 36.6 & 41.8 & 38.2 & 51.4 & 44.0 & 44.8   \\
\bottomrule\bottomrule
\end{tabular}%
}
\end{table*}

\section{Experiments}
\label{sec:experiments}
\subsection{Experiment Settings}
\label{sec:experiments settings}
\noindent\textbf{Evaluated Models.} 
OmniAssisBench requires candidate models to be capable of processing concurrent video and audio streams. Our evaluation suite comprises four closed-source models, \textbf{Gemini-2.5-Pro}~\cite{comanici2025gemini}, \textbf{Gemini-3-Pro}~\cite{deepmind_gemini3pro_2025}, \textbf{Doubao-Seed-2.0-lite}~\cite{seed2026modelcard}, \textbf{MiMo-V2-Omni}~\cite{mimo_v2_omni}, and \textbf{Qwen3.5-Omni-Plus}~\cite{qwen35omniblog}, alongside five open-source models: \textbf{Qwen3-Omni-Instruct (30B)}~\cite{Qwen3-Omni}, \textbf{Qwen2.5-Omni (7B)}~\cite{xu2025qwen2}, \textbf{MiniCPM-o-4.5}~\cite{yu2025minicpm45}, \textbf{MiniCPM-o-2.6}~\cite{yao2024minicpm}, \textbf{Baichuan-Omni-1.5}~\cite{li2025baichuan}, and \textbf{VITA-1.5}~\cite{fu2025vita}. Regarding input preprocessing, all videos are maintained at their native resolution ($\approx$1080p). For models except the Gemini family, videos are sampled at a rate of 1 frame per second (FPS). All other inference parameters adhere to the official recommended configurations for each respective model. 

\noindent\textbf{Context Management.} To accommodate the limited context windows of models during long video inputs, we employ a first-in-first-out strategy, discarding the oldest interaction history when the total input token length exceeds the model's context limit.

\noindent\textbf{Prompting Strategy.} We utilize a unified system prompt designed to elicit behavior characteristic of a helpful video assistant. A critical component is proactive abstention, where the model need to autonomously evaluate whether the visual input necessitates a response. To enforce this, the prompt explicitly instructs the model:

{\footnotesize
\begin{verbatim}
You are a helpful video assistant. For the above 
input video(s), if you decide to output, organize 
your output as if you are directly talking to the 
user. Otherwise, if you decide to keep quiet, 
output exactly "[KEEP QUIET]".
\end{verbatim}
}

\textbf{Scoring Rubric.} 
Model performance is evaluated using an LLM-based automated judge with a structured 5-point scoring rubric. The judge is instantiated with GPT-5 and conducts rubric-based evaluations according to the criteria described below. A detailed list of the scoring rubric can be found in Appendix~\ref{appd: score}. All results reported in the tables throughout this paper have been normalized to a 100-point scale for a more intuitive presentation.

Within this rubric, a score of 60 is defined as the minimum threshold for a valid response. Scores of 60 or higher indicate that the model produces at least a partially meaningful output, while a score of 40 indicates that the model understands the instruction but fails to generate a factually correct response. Scores of 20 or lower correspond to a fundamental failure to comprehend the user's instruction.

In addition to factual correctness, the rubric penalizes unnecessary redundancy to discourage shortcut behaviors such as indiscriminately captioning an entire video to maximize keyword coverage. For proactive-response tasks, scores below 60  reflect insufficient inhibitory control, where the model generates hallucinated or irrelevant content rather than appropriately determining whether a response is warranted.

\subsection{Evaluation Results}
\subsubsection{Overall Performance} 
Tables~\ref{tab:objective_results} and~\ref{interactive_results} present the overall performance of all evaluated models on the Basic and Advanced Interaction Understanding tasks, respectively. The leading closed-source model, \textbf{Gemini-3-Pro}, achieves a score of \textbf{66.4}, while the top-performing open-source model, \textbf{Qwen3-Omni-Instruct}, scores \textbf{51.2}. According to our scoring rubric, these results suggest that current MLLMs generally succeed in understanding verbal prompts but struggle with providing accurate and comprehensive answers.

\subsubsection{Basic Tier Performance}
For Basic Interactive Understanding tasks, the primary challenges stem from viual perceiving. Models usually struggle to accurately perceive non-salient objects and distinguishing target objects from visually similar distractors. Moreover, even a modest increase in the number of target objects (\eg, more than 10) results in a decline in performance, indicating limited scalability in complex visual scenes.

Both open-source and proprietary models struggle with following gesture instructions. There might be two reasons. First, from a perception perspective, gestures are inherently abstract and susceptible to visual ambiguity, making them difficult to distinguish from incidental character movements or background visual noise. Second, from a data perspective, current models likely suffer from insufficient exposure to gesture-grounded instructions during training. While gesture prompts are less prevalent than audio commands, they are indispensable for inclusive interaction, particularly for individuals with speech impairments. This underscores an urgent need to address the gap in interpreting subtle, non-verbal visual cues.

\subsubsection{Advanced Tier Performance}

One primary challenge comes from limitation of model context length. With a 32k context window, MiniCPM-o-2.6 can retain approximately 380 seconds of video at 1 fps. Beyond this limit, the model starts to produce nonsensical words. In contrast, the Qwen family encodes each video frame into a larger number of tokens, resulting in a shorter effective memory of only around 80 seconds. These context capacities are insufficient to preserve a complete interaction history for the multi-turn tasks in our benchmark. Without specialized mechanisms for long-term memory, context limitations are likely to become a critical bottleneck for current models. 

Another common failure is that models struggle to delay responses for the correct time, neither they state that the current information is insufficient for answering. In turns where no user prompts are present, models are easily distracted by the speeches in the video and forget the original goal. 

Also, when users predefine multiple similar tasks, especially when user assign names to each task, most models (exclude Gemini-3-Pro) can describe what is happening but struggle to link the descriptions with the corresponding tasks. This may indicates that although models are simply filter for information according to the object names appeared in the user prompts, instead of truly keep tracking the process based on the predefined TODO lists.

\subsubsection{Real World Cases Performance}
For the three Real World Cases, there is a clear gap between closed-source models (average score of 51.2) and open-source models (average score of 34.8). Although Gemini-3-Pro outperformed its previous counterpart by more than 20 points, it still struggles to correctly perceive and track the status of multiple small objects during long-term interaction in the \textit{Handicraft Process Tracking} task, especially when the target objects are moved out of the camera's field of view. In addition to this challenge, most models fail to understand the map and associate it with the real-world environment in the \textit{Blind Assistance} task, and tend to provide unrelated information in the \textit{Meeting Simulation} task instead of attempting to identify the target person. These results indicate that current open-source models are still far from being satisfactory assistants in real world scenarios, and that even the best closed-source models leave substantial room for improvement.

\begin{figure*}[th]
\centering
    \centerline{\includegraphics[width=\linewidth]{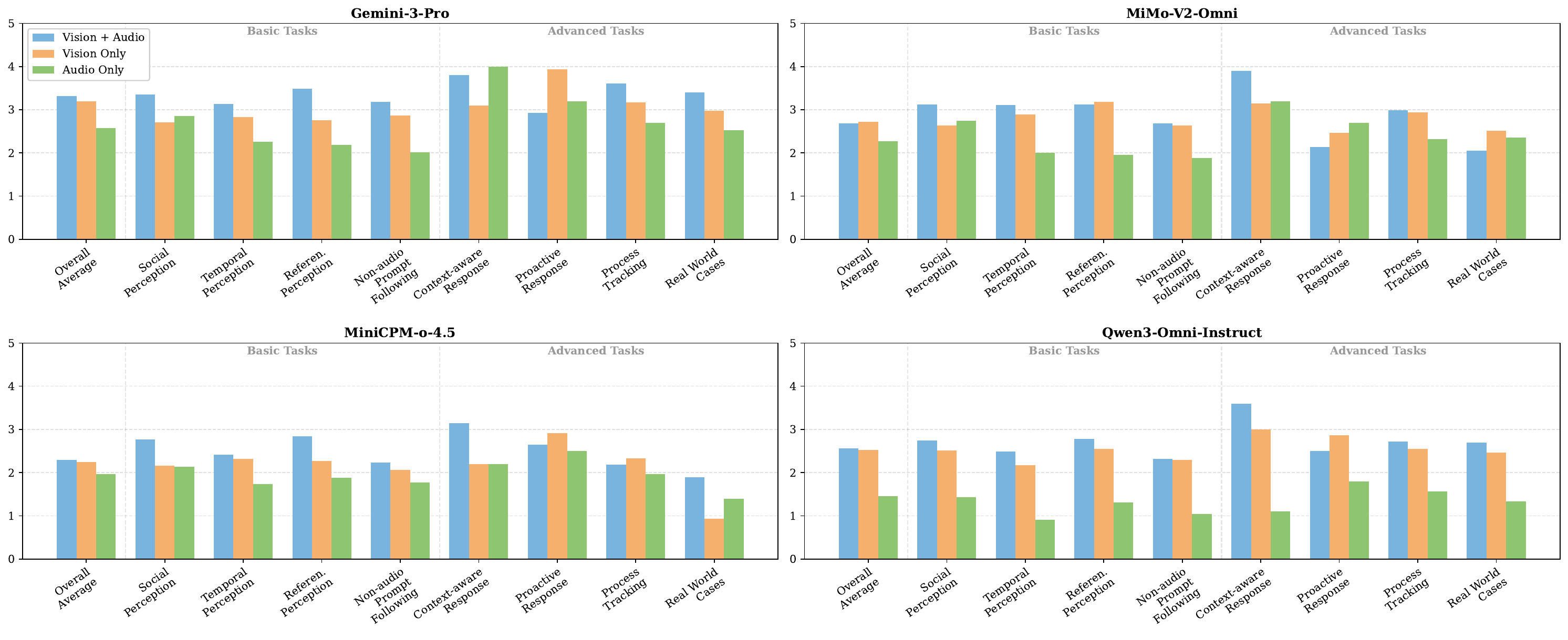}}
    \caption{
      Major-task level performance comparison of four candidate models under original, visual-only, and audio-only input conditions. While removing visual or audio inputs generally degrades performance, models perform better in the Proactive Response task with visual-only inputs, as this prevents them from mistakenly answering background speech instead of the user query. Additionally, Gemini-3 scores higher in the Context-aware Response task under the audio-only setting,  as it avoids formatting penalties triggered by outputting timestamps when video is present.
    }
\label{fig:modality-ablation}
\end{figure*}

\begin{figure*}[th]
\centering
    \centerline{\includegraphics[width=\linewidth]{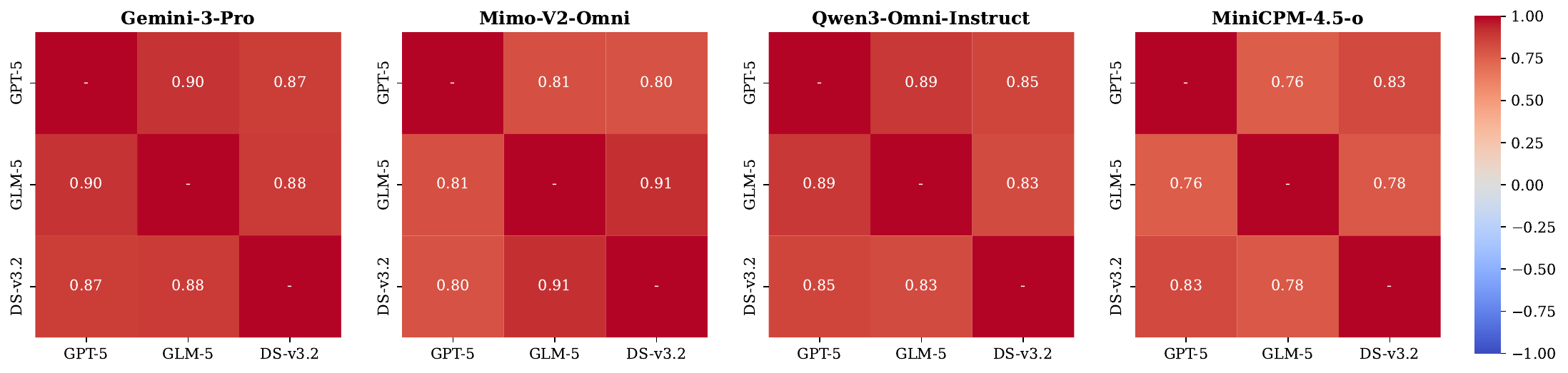}}
    \caption{
      Heat maps of Pearson correlation coefficients between evaluation scores from different judge LLMs across 4 candidate models. DS-v3.2 stands for DeepSeek-v3.2. All coefficient values are above 0.75, indicating that different judge models are likely to agree with each other on whether a candidate's answer is good or not. According to the detailed scores listed in Appendix~\ref{appendix:llm_ablation}, all judge models give similar scores for the candidates.
    }
\label{fig:llm_correlation}
\end{figure*}

\subsection{Ablation Studies}
We conduct four ablations studies. First, we conduct ablation studies on modalities and the consistency of different judge models. We also conduct an ablation study to address the question of balancing the difficulty between single- and multi-turn questions. Also, we study the question of whether down-sampling video frames for longer context can improve model performance.

\begin{figure*}[th]
\centering
    \centerline{\includegraphics[width=\linewidth]{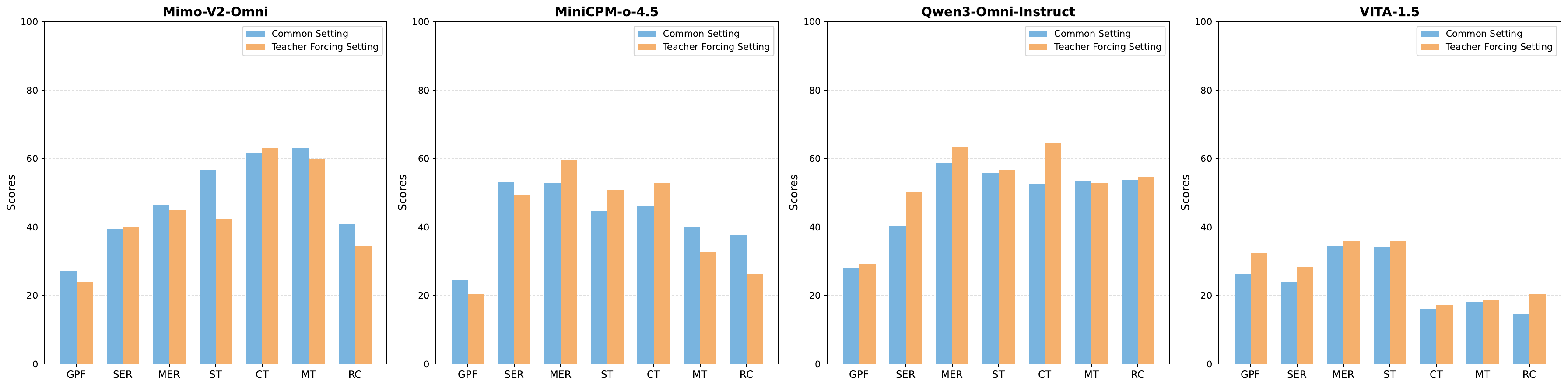}}
    \caption{
    Performance comparison on multi-turn tasks under the common setting and the teacher forcing setting, where historical model responses are replaced by ground truth answers to prevent error accumulation and bridge the difficulty gap between multi- and single-turn tasks. The results indicate that independently evaluating each question in a multi-turn interaction does not always improvements model scores. Abbreviations of task names are defined in Fig.~\ref{fig:data_statistic}. All original 0–5 scores have been normalized to 0–100.
    }
\label{fig:tf-ablation}
\end{figure*}

\begin{figure}[th]
\centering
    \centerline{\includegraphics[width=\linewidth]{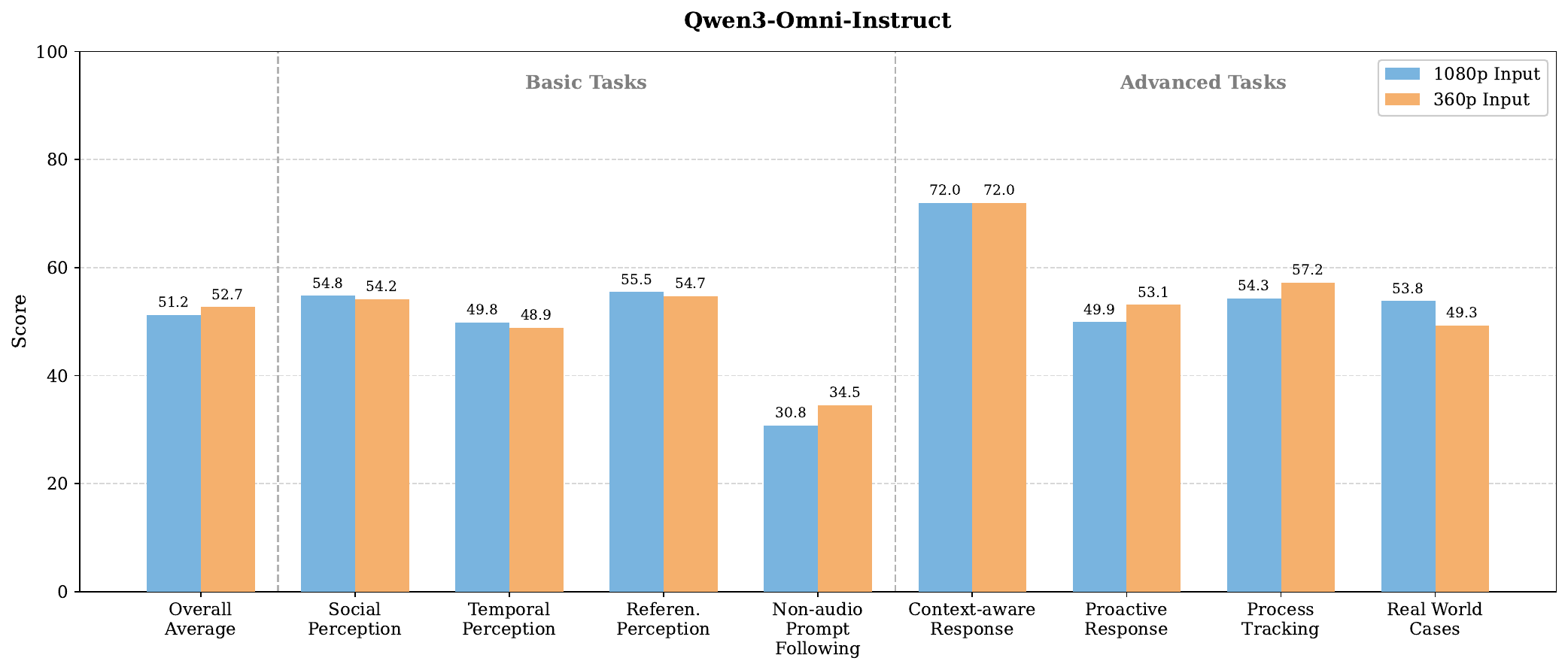}}
    \caption{
    Performance comparison of Qwen3-Omni-Instruct under different input video frame sizes (1080p and 360p). Reducing the input resolution to fit more video frames within the same context length does not significantly improve performance. All original 0–5 scores have been normalized to 0–100.
    }
\label{fig:360p-ablation}
\end{figure}

\subsubsection{Ablation on Input Modalities}
To understand the necessity of the input modalities, we conduct an ablation study by removing either the audio or the visual components from the original test samples. For the audio-only ablation, we simply removed all video frames. For the visual-only ablation, all audios are removed. Consequently, user prompts originally given as speech were converted into text. Prompts given as subtitles remained unchanged. For gesture prompts, the original spoken instructions defining the meaning of the gestures were converted to text, while the gestures themselves remained. Models still need to understand the gestures. We tested four candidate models under these conditions: Gemini-3-Pro, Mimo-V2-Omni, MiniCPM-o-4.5, and Qwen3-Omni-Instruct. The results at the major-task level are summarized in the histogram in Fig.~\ref{fig:modality-ablation}, with detailed scores provided in Appendix~\ref{appendix:llm_ablation}.

The histogram shows that on most tasks, removing either the visual or the audio modality causes a decrease in performance. This drop is especially noticeable for the audio-only inputs, confirming that removing one modality strips away information necessary to correctly answer the questions. However, for the visual-only inputs, the decrease in scores was not very obvious for some tasks. A potential reason is that those models already had relatively low scores on the original inputs. Because they were initially struggling to process the combined audio and video, removing the audio mostly caused them to swap one incorrect answer for a different incorrect answer, resulting in little change to their overall score.

Despite the general drop in performance, we observe an interesting phenomenon: on the "Proactive Response" task, all models actually performed better with visual-only inputs. The proactive response task is a multi-turn task where the user prompt is only given in the first turn. In the following turns, the model receives new multimodal inputs and need to decide on its own whether to respond and what to say. In the original, full-audio inputs, models tend to mistook background speech in the video for new instructions from the user, causing them to forget the original prompt. In the visual-only setting, the absence of distracting audio allowed the models to stay focused on the initial user prompt. This suggests a limitation in current models: they tend to blindly reply to any spoken query they hear, rather than successfully identifying and keeping track of a specific user across multiple turns.

Also, we notice that Gemini-3-Pro performs better with audio-only input on the Context-aware Response task. According to the model replies, this result can be attribute to the model's output format: With video input, model tend to follow an output format of ``timestamp--cue description'', which is punished by our scoring rubric for being less aligned with spoken language. However,  model tend to output colloquial language for audio-only inputs, thus scoring higher.

\subsubsection{Consistence of Different Judge Models}
To understand how the choice of the evaluator affects the results, we conducted an ablation study comparing three different judge LLMs: GPT-5, GLM-5, and DeepSeek-v3.2. In this experiment, each judge model is given the exact same scoring criteria in its prompt and asked to evaluate the outputs of four different candidate models (Gemini3-Pro, Mimo-V2-Omni, MiniCPM-o-4.5 and Qwen3-Omni-Instruct). To measure how closely the judge models agreed with one another, we calculated the Pearson correlation coefficient for every pair of judges evaluating the same candidate. As shown in the four heat maps in Fig.~\ref{fig:llm_correlation}, all correlation values are close to or higher than 0.8. This strong positive correlation indicates a high level of agreement among the three judge models. It shows that, regardless of the specific judge LLM used, the scores can reliably reflect the actual performance of the candidate models. Please refer to Appendix~\ref{appendix:llm_ablation} for detailed scores given by these judge LLMs.

\subsubsection{Teacher Forcing Test Setting for Multi-turn Tasks}
Multi-turn questions are usually considered more difficult than single-turn questions, as errors that accumulated in previous turns may influence model's performance on the following turns. To address this question, we tested the open-sources models under a \textit{teacher forcing} setting: When obtaining the model output at the $T$th turn, we replace all preceding model responses in the $T-1$ turns of interaction history with ground truth answers. This prevents error propagation from previous turns, ensuring that the prediction for the current turn is not affected by historical inaccuracies.

The results comparison of the common setting (without changing the chat history) and the teacher forcing setting are illustrated in Fig.~\ref{fig:tf-ablation}. Although providing a correct interaction history can improve the scores, the improvement is not always obvious. We observe that models quickly adopt to the format of the ground truth answer, which are usually briefer than common model output. However, models still fail to link the ground truth answer with the input videos and take advantages from them.

\subsubsection{Influence of Input Video Resolution}
To address the trad-off between the resolution of input video frames and the total number of frames that can be hold by the maximum context length of the open-source models, we evaluated Qwen3-Omni-Instruct using resized videos where the minimum edge length is fixed to 360 pixels. Under this setting, the model is able to accommodate temporal context of approximately 50 seconds longer than 1080p input.

As shown in Fig.~\ref{fig:360p-ablation}, although holding more video frames in the model context slightly increase the scores on most multi-turn questions, there is no obvious change on the scores across the whole benchmark. On one hand, 50 seconds longer context may still be insufficient for long videos that may last more than 10 minutes, such as videos in the Real-World Cases task. On the other hand, context limitation may not be the only bottleneck that prevent the model from giving valid responses.

%% file: sec/relatedwork.tex
\section{Related Work}
\label{sec:related_work}
\subsection{Advances in Omni MLLMs}
The development of LLMs has evolved from static and discrete processing paradigms~\cite{liu2023llava, zhu2024minigpt}, toward dynamic Omni-LLMs that function as general-purpose multimodal assistants~\cite{pmlr-v235-wu24e,PaLM-E,fei2025on}. These models are capable of concurrently processing video, audio, and text streams, thereby enabling seamless and real-time human-computer interaction. Proprietary models, such as GPT-4o~\cite{openai_gpt4o_2024} and Gemini-3-Pro~\cite{deepmind_gemini3pro_2025}, currently represent the state of the art, demonstrating strong cross-modal reasoning capabilities and highly fluid interactive performance. In parallel, the open-source community has made substantial progress in narrowing the performance gap. Early studies~\cite{li2025baichuan,fu2024vita} primarily focused on aligning multimodal encoders with LLMs to establish basic multimodal interaction capabilities. Building upon this foundation, subsequent works~\cite{fu2025vita, xu2025qwen2} have emphasized latency reduction and deployment efficiency by adopting end-to-end architectures, thereby mitigating the inherent bottlenecks of cascaded systems. The MiniCPM-o family~\cite{yao2024minicpm, yu2025minicpm45} further demonstrates the feasibility of efficient streaming inference on edge devices. More recently, Qwen3-Omni~\cite{Qwen3-Omni} establishes a new baseline by integrating explicit reasoning mechanisms with unified multimodal generation.

\subsection{Video Understanding Benchmarks}
Existing benchmarks in video understanding~\cite{mangalam2023egoschema, li2024mvbench, ning2025video, fu2025video} have primarily focused on evaluating the perception of visual content. However, these benchmarks typically rely on single-turn, multiple-choice questions and restrict their focus exclusively to the visual modality. With the improved abilities of Multimodal LLMs, video understanding benchmarks have begun to include more complex video formats, such as streaming videos, omni-modal videos (understanding both video frames and the corresponding audio), and egocentric videos.
Streaming video understanding benchmarks usually focus on abilities related to temporal perception~\cite{lin2024streaming, zhang2024flash, niu2025ovo}. Omni understanding benchmarks usually focus on the combined perception and alignment of visual and audio modalities~\cite{li2025omnibench, li2025omnivideobench}. Egocentric video understanding benchmarks tend to design tasks similar to those in general video understanding, but test them on egocentric videos~\cite{mangalam2023egoschema, zhou2025egotextvqa}.
In general, although there is a trend for video understanding benchmarks to include more complex video formats that are closer to real-world video chat situations, most of these benchmarks still evaluate abilities independently by design. Also, most benchmarks adopt the QA pair format of a video followed by textual questions. These benchmarks are designed to explore the boundaries of fundamental model abilities, which may not directly reflect how models perform in everyday use.
\subsection{Interactive Understanding Benchmarks}
Compared to video understanding, interactive understanding evaluation is relatively under-explored. ProactiveVideoQA~\cite{wang2025proactivevideoqa} focuses only on proactive abilities. OmniMMI~\cite{wang2025omnimmi} classifies interactive understanding tasks into streaming video understanding and proactive reasoning, which tends to mix interactive tasks with perception tasks. For example, the "speaker identification" task is classified as a proactive reasoning task, even though it is less related to requiring the model to decide "when to answer". EgoLife~\cite{yang2025egolife} evaluates tasks such as tracking social relationships and personal habits in long-duration videos, yet it contains only single-turn multiple-choice questions in text. LifeEval~\cite{gao2026lifeeval} focuses more on testing whether the model can actively associate visual inputs with user intentions, yet it mainly contains short videos under 1 minute. While these pioneering works have explored interactive understanding evaluation from different perspectives, a comprehensive evaluation pipeline has yet to be established.

%% file: sec/conclusion.tex
\section{Conclusion}
This paper introduces OmniAssistBench, a comprehensive benchmark for evaluating Omni-LLMs as interactive assistants. Under interactive scenarios, the model's output directly influence the action of the user, resulting in divergence of interaction paths which static evaluation dataset cannot adapt to. We propose an annotation method to address this problem. We summarize the interaction path based on the source video as prior knowledge. After fixing the path, we are able to reverse-engineering the source video to design test questions and simulate test data by video editing. Based on this annotation pipeline, we build OmniAssistBench and test popular Omni-LLMs. Experiment results indicate the limitations of current models, particularly in gesture instruction following, long-context memory, capacity of delaying response and maintaining information cross turns.

%% file: sec/appendix.tex
\section*{Appendix}
\section{Scoring Rubric}

\label{appd: score}
Table~\ref{tab:scoring_rubric} demonstrates the detailed scoring rubric of our benchmark. The evaluation is based on three dimensions: Accuracy and Completeness, Redundancy, and Adherence to Silence Constraints. These rules are provided to the judgment LLM during evaluation.

\begin{table*}[hb]
\centering
\small 
\setlength{\doublerulesep}{2pt}
\caption{The Scoring Rubric used in OmniAssistBench. The evaluation is based on three dimensions: Accuracy and Completeness, Redundancy, and Adherence to Silence Constraints. These rules are provided to the judgment LLM as prompts during evaluation.}
\label{tab:scoring_rubric}
\renewcommand{\arraystretch}{1.1}

\begin{tabular}{c|>{\vspace*{3pt}}m{5cm}<{\vspace*{3pt}}|>{\vspace*{3pt}}m{5.0cm}<{\vspace*{3pt}}|>{\vspace*{3pt}}m{5.0cm}<{\vspace*{3pt}}}
\toprule\toprule
\textbf{Score} & 
\textbf{Accuracy \& Completeness} & 
\textbf{Redundancy} & 
\textbf{Response to Silence} \newline \textit{(GT = ``KEEP QUIET'')} \\ 
\midrule

\textbf{5} & 
\textbf{Perfect Execution.} The model correctly comprehends the intent and semantically matches the Ground Truth (GT). It covers \textit{all} Key Points. & 
\textbf{None.} Contains almost no redundant information; only minor, contextually necessary elaborations are allowed. & 
\textbf{Exact Match.} Correctly judges that silence is needed and outputs the exact required token ``[KEEP QUIET]''. \\ 
\midrule

\textbf{4} & 
\textbf{Minor Defects.} Comprehends the intent and covers all Key Points, but the response contains minor factual errors or lacks precision in description. & 
\textbf{Low.} Contains minor redundant information, provided it remains relevant to the core subject. & 
\textbf{Descriptive.} Describes the state of inaction but fails to follow the specific instruction to output the exact silence token. \\ 
\midrule

\textbf{3} & 
\textbf{Partial Success.} Comprehends the intent but misses at least one Key Point. The answer is directionally correct but factually imperfect. & 
\textbf{Moderate.} Contains information not present in the GT, but it remains relevant to the general context of the video. & 
\textbf{Evasive.} States that there is insufficient information to answer. \\ 
\midrule

\textbf{2} & 
\textbf{Contradiction.} Comprehends the intent but provides a clearly wrong answer (e.g., opposite polarity, wrong object) or directly contradicts the GT. & 
\textbf{High.} Contains significant irrelevant information, obvious hallucinations, or details unrelated to the query. & 
\textbf{Hallucinated Response.} Fails to understand that current turn is not a proper time for response. Gives hallucinated answers to the questions instead. \\
\midrule

\textbf{1} & 
\textbf{Loose Relation.} Fails to comprehend the specific question. The response is only loosely related to the topic. & 
\multicolumn{1}{c|}{---} & 
\textbf{Loose Relation.} Provides a response that is tangentially related to the context but completely ignores the silence constraint. \\ 
\midrule

\textbf{0} & 
\textbf{Failure.} Complete failure to comprehend the input. The response is entirely unrelated to the question, instruction, or video content. & 
\multicolumn{1}{c|}{---} & 
\textbf{Failure.} Output is entirely unrelated to the question or context. \\ 
\bottomrule\bottomrule
\end{tabular}
\end{table*}

\clearpage

\section{Detailed Task Design}
\label{appd:detailed_task}
Table~\ref{tab:detailed_adv_task} and Table~\ref{tab:detailed_basic_task} list all tasks contained in OmniAssistBench. The italic words briefly explain the key properties which are used for filtering videos during the data collection process.

\begin{table*}[htb]
\centering
\small
\setlength{\doublerulesep}{2pt}
\caption{Detailed table of all Advanced Understanding sub-tasks. \textit{Italic} sentences are explanations of each task.} 
\label{tab:detailed_adv_task}
\renewcommand{\arraystretch}{1.1} 
\begin{tabular}{c|c|c|>{\vspace*{3pt}}m{10.0cm}<{\vspace*{3pt}}|c}
\toprule\toprule
\textbf{Major Task} & \textbf{Abbr.} & \textbf{Sub Task} & \centering \textbf{Description} & \textbf{Turns} \\ 
\midrule

\multirow{2}{*}[1.5em]{\rotatebox[origin=c]{90}{\textbf{Proactive Response}}} & 
\textbf{SER} & \textbf{\begin{tabular}[c]{@{}c@{}}Single Event-\\ triggered \\ Response\end{tabular}} & 
``When the next taxi comes, tell me its license plate number.'' \newline \textit{Models need to wait until the target object appear, then give the answer. We do not explicitly ask the model whether to respond at the end of each turn. Instead, the user prompt is provided only in the first turn, followed by raw video clips in subsequent turns. The model must autonomously determine the appropriate timing for a response.} & 
\textbf{\begin{tabular}[c]{@{}c@{}}4 turns\\ in avg.\end{tabular}} \\ \cmidrule{2-5} 

& \textbf{MER} & \textbf{\begin{tabular}[c]{@{}c@{}}Multi Event-\\ triggered \\ Response\end{tabular}} & 
``Whenever the professor flips to a new slide, tell me its title.'' \newline \textit{Models need to wait until the target object appear. The target event may appear several times, but the user prompt appears only once in the first turn. } & 
\textbf{\begin{tabular}[c]{@{}c@{}}4 turns\\ in avg.\end{tabular}} \\ 
\midrule

\multirow{3}{*}[-3em]{\rotatebox[origin=c]{90}{\textbf{Process Tracking}}} & 
\textbf{ST} & \textbf{\begin{tabular}[c]{@{}c@{}}Step \\ Tracking\end{tabular}} & 
``I want to make this dish strictly following this example video. What should I do next?'' \newline \textit{To avoid ambiguity, we provide a reference (manual images or demonstration videos) at the beginning. User might ask for the next step after completing multiple steps or while halfway through one, preventing shortcut responses that merely recite the next sentence} & 
\textbf{\begin{tabular}[c]{@{}c@{}}6 turns\\ in avg.\end{tabular}} \\ \cmidrule{2-5} 

& \textbf{CT} & \textbf{\begin{tabular}[c]{@{}c@{}}Checklist \\ Tracking\end{tabular}} & 
``This is my shopping list. Which items have not been bought yet?'' \newline \textit{A checklist is provided at the beginning as an image rendered from text. Models need to remember which objects have appeared. Objects may appear in a different order compared with the list.} & 
\textbf{\begin{tabular}[c]{@{}c@{}}4 turns\\ in avg.\end{tabular}} \\ \cmidrule{2-5} 

& \textbf{MT} & \textbf{\begin{tabular}[c]{@{}c@{}}Multitask \\Tracking\end{tabular}} & 
``Here is a TODO list and brief steps of how to finish each task. What is the progress of everything now?'' \newline \textit{The user constantly switch between multiple tasks. There are strict sequential steps within a task, but no dependency between tasks. Models need to track the status of each task, and remind the user whenever required.} & 
\textbf{\begin{tabular}[c]{@{}c@{}}4 turns\\ in avg.\end{tabular}} \\ 
\midrule

\rotatebox[origin=c]{90}{\textbf{\begin{tabular}[c]{@{}c@{}}Context-\\Aware\\ Response\end{tabular}}} & 
\textbf{CR} & \textbf{\begin{tabular}[c]{@{}c@{}}Context-\\Aware\\ Response\end{tabular}} & 
``I'm on the FFmpeg download page. How can I download FFmpeg for this computer?'' \newline \textit{It can be inferred from the video that the user is using a Windows machine. We prefer the model to tailor its suggestions exclusively to the Windows platform, instead of providing generic suggestions covering all major operating systems.} & 
\textbf{\begin{tabular}[c]{@{}c@{}}Single\\ turn\end{tabular}} \\ 
\bottomrule\bottomrule
\end{tabular}
\end{table*}

\begin{table*}[tbp]
\centering
\small
\setlength{\doublerulesep}{2pt}
\caption{Detailed table of all Basic Understanding sub-tasks. \textit{Italic} sentences are explanations of the task.} 
\label{tab:detailed_basic_task}
\renewcommand{\arraystretch}{1.1} 
\begin{tabular}{c|c|c|>{\vspace*{3pt}}m{10.0cm}<{\vspace*{3pt}}|c}
\toprule\toprule

\textbf{\begin{tabular}[c]{@{}c@{}}Major\\Task\end{tabular}} & \textbf{Abbr.} & \textbf{Sub Task} & \centering \textbf{Description} & \textbf{Turns} \\ 
\midrule

\multirow{3}{*}{\rotatebox[origin=c]{90}{\textbf{Social Perception}}} & 
\textbf{II} & \textbf{\begin{tabular}[c]{@{}c@{}}Identity\\Identification\end{tabular}} & 
``Who said that sentence? Who performed that action?'' \newline \textit{Models need to associate multiple actions or speeches with the same person, instead of simply giving answers like ``the man in red'' which can be derived from visual information directly without understanding actual identity.}& 
\textbf{\begin{tabular}[c]{@{}c@{}}Single\\ turn\end{tabular}} \\ \cmidrule{2-5} 

& \textbf{AI} & \textbf{\begin{tabular}[c]{@{}c@{}}Addressee\\Identification\end{tabular}} & 
``Who are they talking to?'' \newline \textit{There are always multiple people in the video apart from the speaker and the addressee. Models may need to check the responses of the people to understand who is the addressee.} & 
\textbf{\begin{tabular}[c]{@{}c@{}}Single\\ turn\end{tabular}} \\ \cmidrule{2-5} 

& \textbf{CE} & \textbf{\begin{tabular}[c]{@{}c@{}}Complex\\Emotion\\ Understanding\end{tabular}} & 
``What emotion or attitude is the person expressing?'' \newline \textit{We only keep complex samples where one modality (e.g., expression or dialogue) alone cannot lead to the correct answer.} & 
\textbf{\begin{tabular}[c]{@{}c@{}}Single\\ turn\end{tabular}} \\ 
\midrule

\multirow{3}{*}{\rotatebox[origin=c]{90}{\textbf{Temporal Perception}}} & 
\textbf{ER} & \textbf{Event Retrieval} & 
``When the main person A said xxx, what color clothes were the person passing behind them wearing?'' \newline \textit{The model needs to establish temporal correlations between the main and non-salient objects or details.} & 
\textbf{\begin{tabular}[c]{@{}c@{}}Single\\ turn\end{tabular}} \\ \cmidrule{2-5} 

& \textbf{AO} & \textbf{\begin{tabular}[c]{@{}c@{}}Appearance\\Order\\ Perception\end{tabular}} & 
``Which colors of clothes did the blogger try on in sequence?'' \newline \textit{Model may need to distinguish the target object from similar ones (e.g., for example, the blogger may have picked up a lot of clothes but only tried on a part of them).} & 
\textbf{\begin{tabular}[c]{@{}c@{}}Single\\ turn\end{tabular}} \\ \cmidrule{2-5} 

& \textbf{DC} & \textbf{\begin{tabular}[c]{@{}c@{}}Dynamic\\Counting\end{tabular}} & 
``How many standard pull-ups did the person perform?'' \newline \textit{Models need to count a large number of instances ($>10$) or filter for specific objects among similar ones.} & 
\textbf{\begin{tabular}[c]{@{}c@{}}Single\\ turn\end{tabular}} \\ 
\midrule

\multirow{2}{*}[2em]{\rotatebox[origin=c]{90}{\textbf{Referential Perception}}} & 
\textbf{AR} & \textbf{\begin{tabular}[c]{@{}c@{}}Action\\Reference\\ Perception\end{tabular}} & 
``Which student is the teacher pointing at? Which file is the mouse point at on the screen?'' \newline \textit{There are always multiple objects similar to the target. In some cases, the referred object cannot be decided by only looking at the ``pointing'' action. Models may need to look for other cues (e.g., there may be multiple students in the direction where the teacher is pointing at, but only the target student stand up after being pointed).} & 
\textbf{\begin{tabular}[c]{@{}c@{}}Single\\ turn\end{tabular}} \\ \cmidrule{2-5} 

& \textbf{LR} & \textbf{\begin{tabular}[c]{@{}c@{}}Linguistic\\Reference\\ Perception\end{tabular}} & 
``What is the title of the second book on the right?'' \newline  \textit{There are always multiple objects similar to the target. The target object may not be the main object of the video.}&
\textbf{\begin{tabular}[c]{@{}c@{}}Single\\ turn\end{tabular}} \\ 
\midrule

\multirow{2}{*}[0.5em]{\rotatebox[origin=c]{90}{\textbf{\begin{tabular}[c]{@{}c@{}}Non-audio\\Prompt Following\end{tabular}}}} & 
\textbf{GPF} & \textbf{\begin{tabular}[c]{@{}c@{}}Gesture-based\\ Prompt \\Following\end{tabular}} & ``This gesture represents describing the cloth wore by the speaker.''\newline
\textit{4 gestures are defined at the beginning of the video, then these gestures are embedded using a picture-in-picture layout as prompts. The gestures as prompts may be made by a different person. Not all pre-defined gestures necessarily appear as prompts, nor do they strictly follow the defined order} & 
\textbf{\begin{tabular}[c]{@{}c@{}}4 turns\\ in avg.\end{tabular}} \\ \cmidrule{2-5} 

& \textbf{OPF} & \textbf{\begin{tabular}[c]{@{}c@{}}OCR-based\\ Prompt\\Following\end{tabular}} & ``What is the name of this calculation rule?''\newline 
\textit{User prompts are embedded as on-screen subtitles or video clips of hand writing using a picture-in-picture layout. There are always other English characters in the video apart from the prompt.} & 
\textbf{\begin{tabular}[c]{@{}c@{}}Single\\ turn\end{tabular}} \\ 
\bottomrule\bottomrule
\end{tabular}
\end{table*}

\clearpage
\section{Consistence of Different Judge Models}
\label{appendix:llm_ablation}
In this section, we provide the detailed scoring data for the ablation study on judge LLMs, which expands upon the heat maps presented in the main text. Table~\ref{table:ablation_llm_basic} and Table~\ref{table:ablation_llm_adv} display the exact evaluation scores given by each of the three judge models (GPT-5, GLM-5, and DeepSeek-v3.2) to the four candidate models across all evaluated tasks. 

\begin{table*}[ht]
\centering
\small
\setlength{\doublerulesep}{2pt}
\caption{Detailed evaluation scores of candidate models assessed by three different judge LLMs under the same scoring criteria. \textit{Overall Avg.} denotes the average score across all evaluated tasks, while this table contains results on the Basic Interactive Understanding tasks. All original 0--5 scores have been normalized to a 0--100 scale.}
\label{table:ablation_llm_basic}
\resizebox{\textwidth}{!}{%
\begin{tabular}{c|c|cccc|cccc|ccc|ccc|c}
\toprule\toprule

\multirow{3}{*}{\textbf{Model}} & \multicolumn{1}{c|}{\multirow{3}{*}{\textbf{\begin{tabular}[c]{@{}c@{}}Overall \\ Avg.\end{tabular}}}} &  \multicolumn{15}{c}{\textbf{ Basic Interactive Understanding Tasks}}\\ \cmidrule(lr){3-17}

 & &  \multicolumn{4}{c|}{\textbf{Social Perception}} & \multicolumn{4}{c|}{\textbf{Temporal Perception}} & \multicolumn{3}{c|}{\textbf{Referential Perception}} & \multicolumn{3}{c|}{\textbf{Non-audio Prompt}} & \multicolumn{1}{c}{\multirow{2}{*}{\textbf{Avg.}}} \\
\cmidrule(lr){3-6} \cmidrule(lr){7-10} \cmidrule(lr){11-13} \cmidrule(lr){14-16}
& & \multicolumn{1}{c}{II} & \multicolumn{1}{c}{AI} & \multicolumn{1}{c}{CE} & \multicolumn{1}{c|}{\textbf{Avg.}} & \multicolumn{1}{c}{ER} & \multicolumn{1}{c}{AO} & \multicolumn{1}{c}{DC} & \multicolumn{1}{c|}{\textbf{Avg.}} & \multicolumn{1}{c}{AR} & \multicolumn{1}{c}{LR} & \multicolumn{1}{c|}{\textbf{Avg.}} & \multicolumn{1}{c}{GPF} & \multicolumn{1}{c}{OPF} & \multicolumn{1}{c|}{\textbf{Avg.}} & \\
\midrule
\multicolumn{16}{l}{\textit{\textbf{Proprietary Models}}} \\
\midrule
\rowcolor{gray!10}
\begin{tabular}[c]{@{}r@{}}Gemini-3-Pro\\(GPT-5)\end{tabular}& 66.4 & 71.8 & 62.8 & 66.2 & 67.0 & 61.2 & 74.6 & 52.6 & 62.6 & 72.6 & 66.4 & 69.8 & 55.0 & 66.6 & 57.4 & 63.6 \\
\begin{tabular}[c]{@{}r@{}}Gemini-3-Pro\\(GLM-5)\end{tabular}& 69.0 & 72.8 & 63.8 & 67.6 & 68.2 & 61.2 & 78.6 & 55.0 & 64.6 & 74.6 & 67.2 & 71.2 & 57.4 & 75.8 & 60.8 & 65.8 \\
\begin{tabular}[c]{@{}r@{}}Gemini-3-Pro\\(DeepSeek-v3.2)\end{tabular}& 68.0 & 74.6 & 65.8 & 71.6 & 70.8 & 65.0 & 74.6 & 61.2 & 66.8 & 74.0 & 65.4 & 70.2 & 54.0 & 72.0 & 57.6 & 65.8 \\
\rowcolor{gray!10}
\begin{tabular}[c]{@{}r@{}}MiMo-V2-Omni\\(GPT-5)\end{tabular} & 53.8 & 66.6 & 56.0 & 64.0 & 62.4 & 50.0 & 77.4 & 60.0 & 62.2 & 67.6 & 56.4 & 62.4 & 27.2 & 62.6 & 34.6 & 53.6 \\
\begin{tabular}[c]{@{}r@{}}MiMo-V2-Omni\\(GLM-5)\end{tabular} & 53.4 & 68.6 & 45.0 & 61.6 & 58.8 & 43.8 & 64.2 & 54.6 & 53.8 & 71.6 & 63.6 & 68.0 & 28.6 & 58.6 & 34.8 & 52.2 \\
\begin{tabular}[c]{@{}r@{}}MiMo-V2-Omni\\(DeepSeek-v3.2)\end{tabular}& 53.6 & 67.6 & 45.0 & 59.2 & 57.6 & 55.0 & 65.4 & 57.4 & 59.2 & 70.8 & 62.8 & 67.0 & 25.8 & 54.6 & 31.8 & 51.8 \\
\midrule
\multicolumn{16}{l}{\textit{\textbf{Open-source Models}}} \\
\midrule
\rowcolor{gray!10}
\begin{tabular}[c]{@{}r@{}}MiniCPM-o-4.5\\(GPT-5)\end{tabular} & 46.0 & 62.8 & 53.4 & 50.8 & 55.4 & 37.6 & 65.4 & 43.8 & 48.4 & 60.0 & 52.8 & 56.8 & 24.6 & 24.0 & 24.4 & 44.6 \\
\begin{tabular}[c]{@{}r@{}}MiniCPM-o-4.5\\(GLM-5)\end{tabular}& 45.8 & 61.8 & 49.6 & 55.4 & 55.6 & 55.0 & 70.0 & 45.0 & 56.0 & 62.2 & 52.8 & 58.0 & 16.6 & 20.0 & 17.4 & 44.2 \\
\begin{tabular}[c]{@{}r@{}}MiniCPM-o-4.5\\(DeepSeek-v3.2)\end{tabular} & 44.0 & 61.0 & 47.6 & 57.6 & 55.6 & 41.2 & 73.4 & 50.0 & 54.4 & 60.0 & 54.6 & 57.6 & 13.8 & 21.4 & 15.2 & 43.0 \\
\rowcolor{gray!10}
\begin{tabular}[c]{@{}r@{}}Qwen3-Omni-Instruct\\(GPT-5)\end{tabular} & 51.2 & 59.0 & 50.4 & 54.6 & 54.8 & 52.6 & 56.0 & 41.2 & 49.8 & 54.8 & 56.4 & 55.6 & 28.2 & 41.4 & 30.8 & 46.4 \\
\begin{tabular}[c]{@{}r@{}}Qwen3-Omni-Instruct\\(GLM-5)\end{tabular} & 49.6 & 64.6 & 49.6 & 55.4 & 56.6 & 52.6 & 57.4 & 35.0 & 48.0 & 56.2 & 51.8 & 54.2 & 22.2 & 36.0 & 25.0 & 44.4 \\
\begin{tabular}[c]{@{}r@{}}Qwen3-Omni-Instruct\\(DeepSeek-v3.2)\end{tabular} & 53.8 & 62.8 & 55.2 & 57.6 & 58.6 & 60.0 & 68.0 & 48.8 & 58.8 & 57.0 & 58.2 & 57.6 & 25.2 & 41.4 & 28.4 & 48.8 \\
\bottomrule\bottomrule
\end{tabular}%
}
\end{table*}

\clearpage
\centering
\footnotesize
\setlength{\doublerulesep}{2pt}
\captionof{table}{Detailed evaluation scores of candidate models assessed by three different judge LLMs under the same scoring criteria. \textit{Overall Avg.} denotes the average score across all evaluated tasks, while this table contains results on the Advanced Interactive Understanding tasks and the Real World Cases. All original 0--5 scores have been normalized to a 0--100 scale.}
\label{table:ablation_llm_adv}
\resizebox{\textwidth}{!}{%
\begin{tabular}{r|c|c|ccc|cccc|c|cccc}
\toprule\toprule

\multirow{3}{*}{\textbf{Model}} &\multirow{3}{*}{\textbf{\begin{tabular}[c]{@{}c@{}}Overall \\ Avg.\end{tabular}}} &  \multicolumn{9}{c|}{\textbf{Advanced Interactive Understanding Tasks}} &\multicolumn{4}{c}{\multirow{2}{*}{\textbf{Real World Cases}}}
\\ \cmidrule(lr){3-11}

 & & \multirow{2}{*}{\textbf{CR}} & \multicolumn{3}{c|}{\textbf{Proactive Response}} & \multicolumn{4}{c|}{\textbf{Process Tracking}} & \multirow{2}{*}{\textbf{Avg.}} & & & &  \\
 \cmidrule(lr){4-6} \cmidrule(lr){7-10} \cmidrule(lr){12-15} 
 & & & SER & MER & \textbf{Avg.} & ST & CT & MT & \textbf{Avg.} & &Ms &Ht & Ba &\textbf{Avg.} \\
\midrule
\multicolumn{12}{l}{\textit{\textbf{Proprietary Models}}} \\
\midrule
\rowcolor{gray!10}
\begin{tabular}[c]{@{}r@{}}Gemini-3-Pro\\(GPT-5)\end{tabular}& 66.4 & 76.0 & 67.8 & 48.4 & 58.6 & 71.8 & 75.0 & 70.2 & 72.2 & 68.2 & 76.4 & 65.8 & 65.0 & 68.0  \\
\begin{tabular}[c]{@{}r@{}}Gemini-3-Pro\\(GLM-5)\end{tabular} & 69.0 & 76.8 & 49.0 & 69.6 & 59.8 & 73.8 & 82.2 & 76.2 & 76.6 & 71.0 & 78.2 & 58.6 & 67.0 & 67.2 \\
\begin{tabular}[c]{@{}r@{}}Gemini-3-Pro\\(DeepSeek-v3.2)\end{tabular} & 68.0 & 79.0 & 43.8 & 65.2 & 55.2 & 73.4 & 81.0 & 74.8 & 75.8 & 69.0 & 80.0 & 61.4 & 69.0 & 69.4 \\
\rowcolor{gray!10}
\begin{tabular}[c]{@{}r@{}}MiMo-V2-Omni\\(GPT-5)\end{tabular} & 53.8 & 78.0 & 39.4 & 46.6 & 42.8 & 56.8 & 61.6 & 63.0 & 59.8 & 55.2 & 52.6 & 27.2 & 44.4 & 41.0 \\
\begin{tabular}[c]{@{}r@{}}MiMo-V2-Omni\\(GLM-5)\end{tabular} & 53.4 & 72.0 & 46.2 & 51.8 & 49.0 & 58.8 & 55.0 & 57.0 & 57.2 & 55.4 & 52.6 & 23.6 & 43.4 & 39.4 \\
\begin{tabular}[c]{@{}r@{}}MiMo-V2-Omni\\(DeepSeek-v3.2)\end{tabular} & 53.6 & 76.0 & 43.8 & 48.2 & 46.0 & 61.2 & 56.0 & 58.8 & 59.2 & 55.8 & 55.0 & 29.0 & 50.0 & 44.8 \\
\midrule
\multicolumn{12}{l}{\textit{\textbf{Open-source Models}}} \\
\midrule
\rowcolor{gray!10}
\begin{tabular}[c]{@{}r@{}}MiniCPM-o-4.5\\(GPT-5)\end{tabular} & 46.0  & 63.0 & 53.2 & 53.0 & 53.0 & 44.6 & 46.0 & 40.2 & 43.8 & 47.8 & 5.4 & 50.0 & 47.0 & 37.8  \\
\begin{tabular}[c]{@{}r@{}}MiniCPM-o-4.5\\(GLM-5)\end{tabular} & 45.8 & 63.0 & 43.4 & 53.8 & 48.8 & 47.6 & 55.2 & 40.6 & 47.6 & 48.8 & 23.6 & 8.6 & 47.8 & 28.8 \\
\begin{tabular}[c]{@{}r@{}}MiniCPM-o-4.5\\(DeepSeek-v3.2)\end{tabular} & 44.0 & 63.0 & 40.6 & 49.6 & 45.2 & 52.2 & 41.6 & 39.8 & 46.0 & 46.6 & 11.0 & 8.6 & 49.0 & 27.2 \\
\rowcolor{gray!10} 
\begin{tabular}[c]{@{}r@{}}Qwen3-Omni-Instruct\\(GPT-5)\end{tabular}  & 51.2 & 72.0 & 40.4 & 58.8 & 50.0 & 55.8 & 52.6 & 53.6 & 54.4 & 53.8 & 67.2 & 40.0 & 56.0 & 53.8 \\
\begin{tabular}[c]{@{}r@{}}Qwen3-Omni-Instruct\\(GLM-5)\end{tabular} & 49.6 & 67.0 & 36.6 & 60.0 & 48.6 & 57.8 & 51.6 & 50.8 & 54.2 & 53.0 & 48.0 & 35.8 & 52.0 & 46.0 \\
\begin{tabular}[c]{@{}r@{}}Qwen3-Omni-Instruct\\(DeepSeek-v3.2)\end{tabular} & 53.8 & 75.0 & 36.8 & 58.6 & 48.2 & 62.6 & 59.4 & 55.0 & 59.6 & 56.6 & 69.0 & 44.2 & 58.0 & 56.4 \\
\bottomrule\bottomrule
\end{tabular}%
}

%% file: main.bib
@String(CVPR= {IEEE Conf. Comput. Vis. Pattern Recog.})

@String(ICLR = {Int. Conf. Learn. Represent.})

@String(CVM = {Computational Visual Media})

@String(CVPR  = {CVPR})

@String(ICLR  = {ICLR})

@inproceedings{liu2023llava,
 author = {Liu, Haotian and Li, Chunyuan and Wu, Qingyang and Lee, Yong Jae},
 booktitle = {NeurIPS},
 title = {Visual Instruction Tuning},
 year = {2023}
}

@inproceedings{
    zhu2024minigpt,
    title={Mini{GPT}-4: Enhancing Vision-Language Understanding with Advanced Large Language Models},
    author={Deyao Zhu and Jun Chen and Xiaoqian Shen and Xiang Li and Mohamed Elhoseiny},
    booktitle={ICLR},
    year={2024},
}

@article{Qwen3-Omni,
  title={Qwen3-Omni Technical Report},
  author={Jin Xu and Zhifang Guo and Hangrui Hu and Yunfei Chu and Xiong Wang and Jinzheng He and Yuxuan Wang and Xian Shi and Ting He and Xinfa Zhu and Yuanjun Lv and Yongqi Wang and Dake Guo and He Wang and Linhan Ma and Pei Zhang and Xinyu Zhang and Hongkun Hao and Zishan Guo and Baosong Yang and Bin Zhang and Ziyang Ma and Xipin Wei and Shuai Bai and Keqin Chen and Xuejing Liu and Peng Wang and Mingkun Yang and Dayiheng Liu and Xingzhang Ren and Bo Zheng and Rui Men and Fan Zhou and Bowen Yu and Jianxin Yang and Le Yu and Jingren Zhou and Junyang Lin},
  journal={arXiv preprint arXiv:2509.17765},
  year={2025}
}

@article{li2025baichuan,
  title={Baichuan-Omni-1.5 Technical Report},
  author={Li, Yadong and Liu, Jun and Zhang, Tao and Chen, Song and Li, Tianpeng and Li, Zehuan and Liu, Lijun and Ming, Lingfeng and Dong, Guosheng and Pan, Da and others},
  journal={arXiv preprint arXiv:2501.15368},
  year={2025}
}

@inproceedings{
    fu2025vita,
    title={{VITA}-1.5: Towards {GPT}-4o Level Real-Time Vision and Speech Interaction},
    author={Chaoyou Fu and Haojia Lin and Xiong Wang and YiFan Zhang and Yunhang Shen and Xiaoyu Liu and Haoyu Cao and Zuwei Long and Heting Gao and Ke Li and Long MA and Xiawu Zheng and Rongrong Ji and Xing Sun and Caifeng Shan and Ran He},
    booktitle={NeurIPS},
    year={2025},
}

@misc{deepmind_gemini3pro_2025,
  author       = {Google DeepMind},
  title        = {Gemini 3 Pro -- Model Card},
  year         = {2025},
  howpublished = {\url{https://storage.googleapis.com/deepmind-media/Model-Cards/Gemini-3-Pro-Model-Card.pdf}},
  note         = {Model card. Published November 2025. Accessed: 2026-01-23}
}

@inproceedings{li2024mvbench,
  title={Mvbench: A comprehensive multi-modal video understanding benchmark},
  author={Li, Kunchang and Wang, Yali and He, Yinan and Li, Yizhuo and Wang, Yi and Liu, Yi and Wang, Zun and Xu, Jilan and Chen, Guo and Luo, Ping and others},
  booktitle={CVPR},
  year={2024}
}

@inproceedings{fu2025video,
  title={Video-mme: The first-ever comprehensive evaluation benchmark of multi-modal llms in video analysis},
  author={Fu, Chaoyou and Dai, Yuhan and Luo, Yongdong and Li, Lei and Ren, Shuhuai and Zhang, Renrui and Wang, Zihan and Zhou, Chenyu and Shen, Yunhang and Zhang, Mengdan and others},
  booktitle={CVPR},
  year={2025}
}

@inproceedings{
li2025omnibench,
title={OmniBench: Towards The Future of Universal Omni-Language Models},
author={Yizhi Li and Ge Zhang and Yinghao Ma and Ruibin Yuan and King Zhu and Hangyu Guo and Yiming Liang and Jiaheng Liu and Zekun Moore Wang and Jian Yang and Siwei Wu and Xingwei Qu and Jinjie Shi and Xinyue Zhang and Zhenzhu Yang and Yidan WEN and Yanghai Wang and Shihao Li and Zhaoxiang Zhang and Ruibo Liu and Emmanouil Benetos and Wenhao Huang and Chenghua Lin},
booktitle={NeurIPS},
year={2025},
}

@article{lin2024streaming,
  title={StreamingBench: Assessing the Gap for MLLMs to Achieve Streaming Video Understanding},
  author={Junming Lin and Zheng Fang and Chi Chen and Zihao Wan and Fuwen Luo and Peng Li and Yang Liu and Maosong Sun},
  journal={arXiv preprint arXiv:2411.03628},
  year={2024}
}

@inproceedings{wang2025omnimmi,
  title={OmniMMI: A Comprehensive Multi-modal Interaction Benchmark in Streaming Video Contexts},
  author={Wang, Yuxuan and Wang, Yueqian and Chen, Bo and Wu, Tong and Zhao, Dongyan and Zheng, Zilong},
  booktitle={CVPR},
  year={2025}
}

@article{wang2025proactivevideoqa,
  title={Proactivevideoqa: A comprehensive benchmark evaluating proactive interactions in video large language models},
  author={Wang, Yueqian and Meng, Xiaojun and Wang, Yifan and Zhang, Huishuai and Zhao, Dongyan},
  journal={arXiv preprint arXiv:2507.09313},
  year={2025}
}

@misc{openai_gpt4o_2024,
  author       = {{OpenAI}},
  title        = {Hello GPT-4o},
  howpublished = {\url{https://openai.com/index/hello-gpt-4o/}},
  month        = may,
  day          = {13},
  year         = {2024},
  note         = {Accessed: 25 Jan 2026}
}

@article{fu2024vita,
  title={Vita: Towards open-source interactive omni multimodal llm},
  author={Fu, Chaoyou and Lin, Haojia and Long, Zuwei and Shen, Yunhang and Dai, Yuhang and Zhao, Meng and Zhang, Yi-Fan and Dong, Shaoqi and Li, Yangze and Wang, Xiong and others},
  journal={arXiv preprint arXiv:2408.05211},
  year={2024}
}

@article{xu2025qwen2,
  title={Qwen2. 5-omni technical report},
  author={Xu, Jin and Guo, Zhifang and He, Jinzheng and Hu, Hangrui and He, Ting and Bai, Shuai and Chen, Keqin and Wang, Jialin and Fan, Yang and Dang, Kai and others},
  journal={arXiv preprint arXiv:2503.20215},
  year={2025}
}

@article{yao2024minicpm,
  title={MiniCPM-V: A GPT-4V Level MLLM on Your Phone},
  author={Yao, Yuan and Yu, Tianyu and Zhang, Ao and Wang, Chongyi and Cui, Junbo and Zhu, Hongji and Cai, Tianchi and Li, Haoyu and Zhao, Weilin and He, Zhihui and others},
  journal={arXiv preprint arXiv:2408.01800},
  year={2024}
}

@inproceedings{
mangalam2023egoschema,
title={EgoSchema: A Diagnostic Benchmark for Very Long-form Video Language Understanding},
author={Karttikeya Mangalam and Raiymbek Akshulakov and Jitendra Malik},
booktitle={NeurIPS},
year={2023},
}

@inproceedings{PaLM-E,
author = {Driess, Danny and Xia, Fei and Sajjadi, Mehdi S. M. and Lynch, Corey and Chowdhery, Aakanksha and Ichter, Brian and Wahid, Ayzaan and Tompson, Jonathan and Vuong, Quan and Yu, Tianhe and Huang, Wenlong and Chebotar, Yevgen and Sermanet, Pierre and Duckworth, Daniel and Levine, Sergey and Vanhoucke, Vincent and Hausman, Karol and Toussaint, Marc and Greff, Klaus and Zeng, Andy and Mordatch, Igor and Florence, Pete},
title = {PaLM-E: an embodied multimodal language model},
year = {2023},
booktitle = {ICML},
}

@InProceedings{pmlr-v235-wu24e,
  title = 	 {{NE}x{T}-{GPT}: Any-to-Any Multimodal {LLM}},
  author =       {Wu, Shengqiong and Fei, Hao and Qu, Leigang and Ji, Wei and Chua, Tat-Seng},
  booktitle = 	 {ICML},
  year = 	 {2024},
}

@inproceedings{
fei2025on,
title={On Path to Multimodal Generalist: General-Level and General-Bench},
author={Hao Fei and Yuan Zhou and Juncheng Li and Xiangtai Li and Qingshan Xu and Bobo Li and Shengqiong Wu and Yaoting Wang and Junbao Zhou and Jiahao Meng and Qingyu Shi and Zhiyuan Zhou and Liangtao Shi and Minghe Gao and Daoan Zhang and Zhiqi Ge and Siliang Tang and Kaihang Pan and Yaobo Ye and Haobo Yuan and Tao Zhang and Weiming Wu and Tianjie Ju and Zixiang Meng and Shilin Xu and Liyu Jia and Wentao Hu and Meng Luo and Jiebo Luo and Tat-Seng Chua and Shuicheng YAN and Hanwang Zhang},
booktitle={ICML},
year={2025},
}

@article{ning2025video,
  title={Video-bench: A comprehensive benchmark and toolkit for evaluating video-based large language models},
  author={Ning, Munan and Zhu, Bin and Xie, Yujia and Lin, Bin and Cui, Jiaxi and Yuan, Lu and Chen, Dongdong and Yuan, Li},
  journal={CVM},
  year={2025},
}

@inproceedings{niu2025ovo,
  title={OVO-Bench: How Far is Your Video-LLMs from Real-World Online Video Understanding?},
  author={Niu, Junbo and Li, Yifei and Miao, Ziyang and Ge, Chunjiang and Zhou, Yuanhang and He, Qihao and Dong, Xiaoyi and Duan, Haodong and Ding, Shuangrui and Qian, Rui and others},
  booktitle={CVPR},
  year={2025}
}

@article{li2025omnivideobench,
  title={Omnivideobench: Towards audio-visual understanding evaluation for omni mllms},
  author={Li, Caorui and Chen, Yu and Ji, Yiyan and Xu, Jin and Cui, Zhenyu and Li, Shihao and Zhang, Yuanxing and Tang, Jiafu and Song, Zhenghao and Zhang, Dingling and others},
  journal={arXiv preprint arXiv:2510.10689},
  year={2025}
}

@article{zhang2024flash,
  title={Flash-vstream: Memory-based real-time understanding for long video streams},
  author={Zhang, Haoji and Wang, Yiqin and Tang, Yansong and Liu, Yong and Feng, Jiashi and Dai, Jifeng and Jin, Xiaojie},
  journal={arXiv preprint arXiv:2406.08085},
  year={2024}
}

@article{comanici2025gemini,
  title={Gemini 2.5: Pushing the frontier with advanced reasoning, multimodality, long context, and next generation agentic capabilities},
  author={Comanici, Gheorghe and Bieber, Eric and Schaekermann, Mike and Pasupat, Ice and Sachdeva, Noveen and Dhillon, Inderjit and Blistein, Marcel and Ram, Ori and Zhang, Dan and Rosen, Evan and others},
  journal={arXiv preprint arXiv:2507.06261},
  year={2025}
}

@INPROCEEDINGS{sener2022assembly101,
  author={Sener, Fadime and Chatterjee, Dibyadip and Shelepov, Daniel and He, Kun and Singhania, Dipika and Wang, Robert and Yao, Angela},
  booktitle={CVPR}, 
  title={Assembly101: A Large-Scale Multi-View Video Dataset for Understanding Procedural Activities}, 
  year={2022},
}

@inproceedings{grauman2022ego4d,
  title={Ego4D: Around the World in 3,000 Hours of Egocentric Video},
  author={Grauman, Kristen and Westbury, Andrew and Byrne, Eugene and Chavis, Zachary and Furnari, Antonino and Girdhar, Rohit and Hamburger, Jackson and Jiang, Hao and Liu, Miao and Liu, Xingyu and others},
  booktitle={CVPR},
  year={2022}
}

@inproceedings{pan2025basket,
  title={BASKET: A Large-Scale Video Dataset for Fine-Grained Skill Estimation},
  author={Pan, Yulu and Zhang, Ce and Bertasius, Gedas},
  booktitle={CVPR},
  year={2025}
}

@inproceedings{rossetto2025castle,
  title={The castle 2024 dataset: Advancing the art of multimodal understanding},
  author={Rossetto, Luca and Bailer, Werner and Dang-Nguyen, Duc-Tien and Healy, Graham and J{\'o}nsson, Bj{\"o}rn {\TH}{\'o}r and Kongmeesub, Onanong and Le, Hoang-Bao and Rudinac, Stevan and Sch{\"o}ffmann, Klaus and Spiess, Florian and others},
  booktitle={ACM MM},
  year={2025}
}

@inproceedings{tang2019coin,
  title={Coin: A large-scale dataset for comprehensive instructional video analysis},
  author={Tang, Yansong and Ding, Dajun and Rao, Yongming and Zheng, Yu and Zhang, Danyang and Zhao, Lili and Lu, Jiwen and Zhou, Jie},
  booktitle={CVPR},
  year={2019}
}

@inproceedings{poria2019meld,
  title={Meld: A multimodal multi-party dataset for emotion recognition in conversations},
  author={Poria, Soujanya and Hazarika, Devamanyu and Majumder, Navonil and Naik, Gautam and Cambria, Erik and Mihalcea, Rada},
  booktitle={ACL},
  pages={527--536},
  year={2019}
}

@inproceedings{ray-etal-2022-multimodal,
    title = "A Multimodal Corpus for Emotion Recognition in Sarcasm",
    author = "Ray, Anupama  and
      Mishra, Shubham  and
      Nunna, Apoorva  and
      Bhattacharyya, Pushpak",
    booktitle = "LREC",
    year = "2022",
}

@inproceedings{oh2011large,
  title={A large-scale benchmark dataset for event recognition in surveillance video},
  author={Oh, Sangmin and Hoogs, Anthony and Perera, Amitha and Cuntoor, Naresh and Chen, Chia-Chih and Lee, Jong Taek and Mukherjee, Saurajit and Aggarwal, Jake K and Lee, Hyungtae and Davis, Larry and others},
  booktitle={CVPR},
  year={2011}
}

@article{yu2025minicpm45,
  title={Minicpm-v 4.5: Cooking efficient mllms via architecture, data, and training recipe},
  author={Yu, Tianyu and Wang, Zefan and Wang, Chongyi and Huang, Fuwei and Ma, Wenshuo and He, Zhihui and Cai, Tianchi and Chen, Weize and Huang, Yuxiang and Zhao, Yuanqian and others},
  journal={arXiv preprint arXiv:2509.18154},
  year={2025}
}

@misc{mimo_v2_omni,
  author       = {{Xiaomi}}, 
  title        = {{Xiaomi MiMo-V2-Omni: Omni-Modal Agentic Foundation Model that Sees, Understands and Acts}},
  year         = {2026}, 
  url          = {https://platform.xiaomimimo.com/#/docs/news/v2-omni-release},
  note         = {Accessed: Mar. 24, 2026}
}

@inproceedings{yang2025egolife,
  title={EgoLife: Towards Egocentric Life Assistant},
  author={Yang, Jingkang and Liu, Shuai and Guo, Hongming and Dong, Yuhao and Zhang, Xiamengwei and Zhang, Sicheng and Wang, Pengyun and Zhou, Zitang and Xie, Binzhu and Wang, Ziyue and Ouyang, Bei and Lin, Zhengyu and Cominelli, Marco and Cai, Zhongang and Li, Bo and Zhang, Yuanhan and Zhang, Peiyuan and Hong, Fangzhou and Widmer, Joerg and Gringoli, Francesco and Yang, Lei and Liu, Ziwei},
  booktitle={Proceedings of the IEEE/CVF Conference on Computer Vision and Pattern Recognition (CVPR)},
  year={2025}
}

@article{gao2026lifeeval,
  title={LifeEval: A Multimodal Benchmark for Assistive AI in Egocentric Daily Life Tasks},
  author={Gao, Hengjian and Zhang, Kaiwei and Wang, Shibo and Chen, Mingjie and Cao, Qihang and Wang, Xianfeng and Zhu, Yucheng and Min, Xiongkuo and Sun, Wei and Zhu, Dandan and Zhai, Guangtao},
  journal={arXiv preprint arXiv:2603.00490},
  year={2026}
}

@inproceedings{zhou2025egotextvqa,
  title={EgoTextVQA: Towards Egocentric Scene-Text Aware Video Question Answering},
  author={Zhou, Sheng and Xiao, Junbin and Li, Qingyun and Li, Yicong and Yang, Xun and Guo, Dan and Wang, Meng and Chua, Tat-Seng and Yao, Angela},
  booktitle={Proceedings of the IEEE/CVF Conference on Computer Vision and Pattern Recognition (CVPR)},
  year={2025}
}

@misc{qwen35omniblog,
    title = {Qwen3.5-Omni: Scaling Up, Toward Native Omni-Modal AGI},
    url = {https://qwen.ai/blog?id=qwen3.5-omni},
    author = {Qwen Team},
    month = {March},
    year = {2026}
}

@misc{seed2026modelcard,
    title={Seed2.0 Model Card: Towards Intelligence Frontier for Real-World Complexity},
    author={{ByteDance Seed Team}},
    year={2026},
    howpublished={\url{https://seed.bytedance.com/en/seed2}}
}
